\documentclass[11pt]{article}

\usepackage{amssymb}
\usepackage{amsmath,amsthm}
\usepackage{graphicx}
\usepackage{subfigure}

\usepackage{amssymb}
\usepackage{pifont}
\DeclareMathOperator*{\argmin}{arg\,min}
\DeclareMathOperator*{\argmax}{arg\,max}

\usepackage{enumitem}
\usepackage{booktabs}      

\newtheorem{theorem}{Theorem}
\newtheorem{proposition}{Proposition} 
 
\newtheorem{Assumption}{Assumption} 
 
\newtheorem{Definition}{Definition}

\usepackage{multirow} 
\usepackage[table,dvipsnames]{xcolor}
\usepackage{colortbl}

\newcommand*{\belowrulesepcolor}[1]{%
	\noalign{%
		\kern-\belowrulesep
		\begingroup
		\color{#1}%
		\hrule height\belowrulesep
		\endgroup
	}%
}
\newcommand*{\aboverulesepcolor}[1]{%
	\noalign{%
		\begingroup
		\color{#1}%
		\hrule height\aboverulesep
		\endgroup
		\kern-\aboverulesep
	}%
}

\usepackage{times}

\usepackage[numbers]{natbib}

\usepackage{bm}

\usepackage{hyperref}
\usepackage{cleveref}

\usepackage{mathrsfs}

\usepackage{fullpage}
\usepackage[protrusion=true,expansion=true]{microtype}

\usepackage{xcolor,colortbl}
\definecolor{DarkBlue}{RGB}{22,54,93}

\usepackage{algorithm}
\usepackage{algorithmic}

\begin{document}
\title{\huge Network Utility Maximization with Unknown Utility Functions:  A Distributed, Data-Driven Bilevel Optimization Approach}

\author
{
Kaiyi Ji$^1$ and Lei Ying$^2$
\vspace{0.1cm}\\$^1:$ Department of CSE, University at Buffalo
\vspace{0.1cm}\\$^2:$ Department of EECS, University of Michigan, Ann Arbor
}
\maketitle
\begin{abstract}
Fair resource allocation is one of the most important topics in communication networks. Existing solutions almost exclusively assume each user utility function is known and concave. This paper seeks to answer the following question: how to allocate resources when utility functions are unknown, even to the users? This answer has become increasingly important in the next-generation AI-aware communication networks where the user utilities are complex and their closed-forms are hard to obtain. In this paper, we provide a new solution using a distributed and data-driven bilevel optimization approach, where the lower level is a distributed network utility maximization (NUM) algorithm with concave surrogate utility functions, and the upper level is a data-driven learning algorithm to find the best surrogate utility functions that maximize the sum of true network utility. The proposed algorithm learns from data samples (utility values or gradient values) to autotune the surrogate utility functions to maximize the true network utility, so works for unknown utility functions. For the general network, we establish the nonasymptotic convergence rate of the proposed algorithm with nonconcave utility functions. The simulations validate our theoretical results and demonstrate the great effectiveness of the proposed method in a real-world network.
\end{abstract}
\section{Introduction}
Network utility maximization (NUM) has been studied for decades since the seminal work \cite{kelly1998rate} and has been the central analytical framework for the design of fair and distributed resource allocation over the communication networks (e.g., the Internet, 6G networks). Its applications span from network congestion control~\citep{kelly1998rate,low1999optimization,palomar2007alternative},  power allocation and routing in wireless networks~\citep{neely2003dynamic,georgiadis2006resource}, load scheduling in cloud computing~\citep{maguluri2012stochastic,ghaderi2016scheduling,ji2018asymptotic}, to video streaming over dynamic networks~\citep{talebi2011quasi,hajiesmaili2012content,gong2014social,li2016data,ye2017charging,fu2022learning}, and etc. A comprehensive introduction of the method and its connections to control theory and convex optimization can be found in \cite{srikant2013communication}.

In the traditional NUM,  each user is associated with a utility function that captures the level of satisfaction with allocated resources (often the assigned data rate),  and distributed NUM solutions and their variations have been implemented as   the congestion control algorithms on the Internet, such as TCP-Reno, and scheduling algorithms for cellular networks, such as Proportional Fair Scheduling. The solutions  maximize the total network utility subject to resource constraints such as channel capacity, average power, etc.  There have been a large body of studies on NUM 
for wired ~\citep{kelly1998rate,low1999optimization,palomar2007alternative,srikant2013communication,Chi_05,lin2006utility,wei2013distributed,beck20141} 
and wireless networks ~\citep{lee2005non,neely2008fairness,ErySri_04,Sto_05,WanKar_05,ChiLowCal_07,weeraddana2011resource,gong2014social,li2016data,ye2017charging,ma2018charging,fu2022learning}.
These existing studies on NUM almost exclusively assume that the utility functions are known to the users and are concave, e.g. the widely used 
$\alpha$-fair utility functions \cite{MoWal_00}, 
However, in many real-world applications, e.g., emerging AI-aware next-generation networks like 6G, the underlying utilities often correlate with user experience, information freshness, diversity, fidelity, job quality, etc, which can be  nonconcave and generally  unknown. Then, an open and challenging question in the field is: 

{\em How to allocate network resources fairly and efficiently  when the utility functions are  unknown and nonconcave? } 

The answer to this question has not been explored well except a few recent attempts~\citep{verma2020stochastic,fu2022learning} using online learning algorithms. For example, \cite{fu2022learning} focused on a stochastic dynamic scenario, and proposed an online policy to gradually learn the utility functions and allocate resources accordingly. However, it still assumes the unknown utility functions are concave and requires a central scheduler. 

In this paper, we consider unknown utility functions and provide a distributed solution  from a new bilevel optimization perspective, where 
the lower-level problem is a standard distributed resource allocation algorithm with parameterized surrogate utility functions such as $\alpha-$fair utility functions, and the upper-level is to fine-tune the surrogate utility functions based on user experiences/feedback. While the solution is based on bilevel optimization, it is very different from existing studies for non-distributed bilevel optimization~\citep{domke2012generic,pedregosa2016hyperparameter,ji2021bilevel,ji2021lower,liu2020generic,ji2022will,ghadimi2018approximation,hong2020two} (see~\cite{liu2021investigating} and \cite{ji2021bilevelo} for a more comprehensive overview) due to the distributed nature of the solution over the communication networks. In addition, these approaches cannot be directly applied here due to the computation of either the Hessian inverse or a product of Hessians of the NUM objective, which requires each node to know the infeasible global network information, which is not practical.  Although several decentralized bilevel optimization methods have been proposed by~\cite{lu2022decentralized,chen2022decentralized,yang2022decentralized,gao2022stochastic,qiu2022diamond}, they consider general bilevel objective functions  without taking the channel capacity, the transmission links and the structured NUM objectives into account, and hence cannot be directly applied to the NUM problems. Then, the main contributions of this paper are summarized below:
\begin{list}{$\bullet$}{\topsep=0.5ex \leftmargin=0.1in \rightmargin=0.1in \itemsep =0.06in}
\item Our first contribution is the design of a distributed bilevel optimization algorithm named DBiNUM, which approximates the Hessian-inverse-vector product of the upper-level gradient using one-step gradient decent. We show that each user under DBiNUM only needs to know the partial network information such as transmission rates and link states of other users on her route, and hence DBiNUM admits a distributed implementation. In addition, DBiNUM does not need to know the true utility functions and only requires user feedback via gradient- or value-based queries. 

\item Theoretically, we prove that the hypergradient estimation error, although large initially, is formed by iteratively decreasing terms with a proper selection of the learning rates. Based on such key derivations, we provide the finite-time convergence rate guarantee for DBiNUM with a general nonconcave upper objective as well as a general network topology.  We further provide a case study for a single-link multi-user network, where we show that when the true user utilities are $\alpha$-fairness functions (but still unknown to the users), DBiNUM converges to the solution as if the utility functions are known. This provides some validation for the proposed bilevel formulation. 

\item In the simulations, we first validate our theoretical result by showing that our bilevel algorithm converges to the standard NUM solutions (total utility, user resources) when the true utility functions are $\alpha$-fair utility functions. In a real-world   
 Abilene network, we demonstrate that our bilevel approach achieves a significantly better network utility than the standard NUM baseline with fixed surrogate utility functions. 
\end{list}

\section{Problem Formulation} 

Consider a communication network with $n$ users (or data flows) and $m$ communication links. Each user is associated with a utility function  $\widetilde U_r(x_r),$ where $x_r$ the transmission rate of user $r.$ Let $\mathcal{L}=\{l_1,l_2,....,l_m\}$ denote all communication links, $c_l$ denote the capacity of link $l,$ $\mathcal{L}_r$ denote all links along the route of user $r$, and $\mathbf{x}=[x_1,...,x_n]^T$ denote the transmission rate vector. The network utility maximization (NUM) problem is to find a resource allocation $\mathbf{x}$ that solves the following optimization problem: 
\begin{align}\label{eq:obj}
&\max_{\mathbf{x}} \sum_{r=1}^n \widetilde U_r(x_r)\\
\hbox{subject to:}&\sum_{r:l\in \mathcal{L}_r} x_r \leq c_l, \quad \text{ for any } l\in\mathcal{L} \nonumber
\\ &\quad x_r \geq 0, \quad\text{ for } r=1,...,n.
\end{align}
Different from existing works on NUM, we consider general utility function $\widetilde U_r(x_r),$ not necessarily concave, and assume it may be unknown to user $r.$ 

\subsection{The Traditional Network Utility Maximization and the Primal Solution}
If  $U_r(\cdot)$ is continuously twice differentiable and {\bf concave}, e.g, $\alpha$-fairness utility function such that $U_r(x_r)=\frac{x_r^{1-\alpha}}{1-\alpha},$ and is known to user $r,$ then the problem in \cref{eq:obj} becomes the traditional NUM problem and has been extensively studied since the seminal work \cite{kelly1998rate}. In particular, 
a  variety of distributed algorithms have been proposed to solve \cref{eq:obj} efficiently with only limited information exchange between the user and the network. Among them, the primal approach penalizes the capacity constraints into the total network utility, and solves the following  alternative  regularized problem.  
\begin{align}\label{eq:primal_obj}
\min_{x_1,...,x_n> 0} \quad \sum_{r=1}^n U_r(x_r) - \sum_{l\in\mathcal{L}}B_l\Big(\sum_{r:l\in\mathcal{L}_r}x_r\Big),
\end{align}
where the regularizer $B_l(\cdot)$ is continuously twice differentiable and $\mu$-strongly-convex, and 
can be regarded as the cost of transmitting the data on link $l$ to penalize the arrival rate for exceeding the link capacity. TCP-Reno for the Internet congestion control is such a primal algorithm.



\subsection{NUM via Bilevel Optimization}



The question we want to answer is how to solve NUM with unknown utility functions and how to solve it in a distributed fashion.
%
%
 We propose a distributed, bilevel solution to this problem. The lower level corresponds to  a standard network resource allocation problem via a primal distributed algorithm as in \cref{eq:primal_obj} with parameterized surrogate utility functions $U_r(x_r;\alpha_r),$ where ${\bm \alpha} \in \mathcal{A}$ are the parameters and the surrogate function is continuously twice differentiable and concave for any given  ${\bm \alpha} \in \mathcal{A}$.  
The upper-level add-on procedure is to fine-tune the user-specified parameters $\alpha_r, r=1,...,n$ to learn the best surrogate utilities $U_r(x_r;\alpha_r),r=1,...,n$ based on the user feedback, e.g., the value-based query $\widetilde U_r(x_r)$ (i.e., how much the user feel satisfied with $x_r$)    
or  the gradient-based query $\nabla \widetilde U_r(x_r)$ (i.e., how fast the user experience increases at $x_r$). 
Mathematically, this problem can be formulated as  
\begin{align}\label{eq:biobj}
\quad &\max_{\bm{\alpha}\in\mathcal{A}}\left[\Psi(\bm{\alpha})=\sum_{r=1}^n \widetilde U_r(x_r^*(\bm{\alpha}))\right] \nonumber
\\&\mathbf{x}^*(\bm{\alpha}) =\argmax_{\bm{x} > 0}\Phi(\mathbf{x};\bm{\alpha}) = \sum_{r=1}^n \big(U_r(x_r;\alpha_r) -\frac{\epsilon x_r^2}{2}\big)-\sum_{l\in\mathcal{L}}B_l\Big(\sum_{i:l\in\mathcal{L}_i}x_i\Big),
\end{align}
where $\mathcal{A}:=\{\bm{\alpha}:\alpha_r\in\mathcal{A}_r, r=1,...,n\}$ is a closed, convex and bounded constraint set.   
Compared with \cref{eq:primal_obj}, we add a small quadratic term $-\frac{\epsilon x_r^2}{2}$ to each surrogate utility function $U_r(x_r;\alpha_r)$ to ensure that the lower-level objective function $\Phi(\bm{x};\bm{\alpha})$ is {\bf strongly-concave} w.r.t.~$\mathbf{x}$. Also note that this extra quadratic term changes the original solution of \cref{eq:primal_obj} up to only an $\epsilon$ level, and hence the solution $\bm{x}^*(\bm{\alpha})$ is still valid.


\section{Algorithm and Main Results}
 
 We first discuss the challenges in solving the bilevel problem \cref{eq:biobj} and then present a distributed bilevel  algorithm. We then provide the main results for the proposed method.  
\subsection{Challenges in Hypergradient Computation over Networks} 
Gradient ascent is a typical method to efficiently solve the bilevel problem in \cref{eq:biobj}.  This process needs to calculate the gradient  $\nabla\Psi(\bm{\alpha})$ (which we refer to the hypergradient) of the upper-level objective function. However, as shown in the following proposition,  this hypergradient contains complicated components due to the nested problem structure. 
\begin{proposition}\label{prop:hyperG}
Hypergradient $\nabla \Psi(\bm{\alpha})$ takes the form of 
\begin{align}\label{eq:hyperGrad}
\nabla \Psi(\bm{\alpha}) = - \nabla_{\bm{\alpha}} \nabla_{\bm{x}} &\Phi(\bm{x}^*;\bm{\alpha}) \big( \nabla^2_{\bm{x}} \Phi(\bm{x}^*;\bm{\alpha}) \big)^{-1}\big[\nabla \widetilde U_1(x_1^*),...,\nabla \widetilde U_n(x_n^*)\big]^T,
\end{align}
where  $\nabla_{\bm{\alpha}} \nabla_{\bm{x}} \Phi(\bm{x}^*;\bm{\alpha})$ is a diagonal matrix whose $i^{th}$ diagonal element is $\nabla_{\alpha}\nabla_{x}U_i(x_i^*;\alpha_i)$, and the $(i,j)^{th}$ element of the Hessian matrix  $\nabla^2_{\bm{x}} \Phi(\bm{x}^*;\bm{\alpha})$ equals to 
\begin{align}\label{eq:hessian_form}
\begin{cases}
\nabla^2_{x} U_i(x_i^*;\alpha_i) -\epsilon- \sum_{l\in\mathcal{L}_i}\nabla^2B_l\big(\sum_{r:l\in\mathcal{L}_r}x_r\big), i=j
 \\ -\sum_{l\in\mathcal{L}_i\cap\mathcal{L}_j}\nabla^2B_l\big(\sum_{r:l\in\mathcal{L}_r}x_r\big),\,i\neq j,
 \end{cases}
\end{align}
where we define $\sum_{l\in\emptyset}(\cdot)=0$ for simplicity. 
\end{proposition}
Note that the Hessian matrix $ \nabla^2_{\bm{x}} \Phi(\bm{x}^*;\bm{\alpha}) $ is invertible because the lower-level function $\Phi(\bm{x};\bm{\alpha})$ is strongly-concave. 
As shown in \Cref{prop:hyperG}, the hypergradient $\nabla\Psi(\bm{\alpha})$ involves the second-order derivatives $\nabla_{\bm{\alpha}} \nabla_{\bm{x}} \Phi(\bm{x}^*;\bm{\alpha})$ and $ \nabla^2_{\bm{x}} \Phi(\bm{x}^*;\bm{\alpha})$ of the lower-level function $\Phi(\bm{x}^*;\bm{\alpha})$. In particular, exactly computing $\nabla\Psi(\bm{\alpha})$ needs to invert the Hessian matrix $ \nabla^2_{\bm{x}} \Phi(\bm{x}^*;\bm{\alpha})$ whose form is taken as in \Cref{prop:hyperG}. However, this inversion is hard to implement in a large communication network because it requires the global network information but each user in reality knows only partial information. In addition, this inversion is computationally infeasible because the matrix dimension can be super large when the network contains millions of users. We next introduce a fast approximation method to tackle these two issues, which 1) allows a distributed implementation in the network and 2) is highly efficient without any Hessian inverse computations.

\subsection{Proposed Distributed Bilevel Algorithm}
In this section, we present a distributed bilevel method for solving the resource allocation problem in \cref{eq:biobj}.
\begin{algorithm}[t]
	\caption{Distributed Bilevel Network Utility Maximization (DBiNUM)} 
	\small
	\label{alg:bio}
	\begin{algorithmic}[1]
		\STATE {\bfseries Input:}  Initialization $ \bm{a}_0\in\mathcal{A}$ and $\mathbf{x}_0>\mathbf{0}$
		\FOR{$k=0,1,...,K$}
\STATE{Lower-level {\bf standard} network maximization procedure with $T_{l}$ time slots:
\begin{itemize}
\item Use {\bf standard} distributed primal algorithm to get $\widehat x_{k,r}\geq0$ satisfying $|\widehat x_{k,r}-x_{k,r}^*|\leq\delta_\Phi$ for each user $r$. 
\end{itemize}
}
\STATE{Information broadcast for upper level with $T_o$ time slots: 
\begin{itemize}
\item All  users release packets with information $\widehat x_{k,r}$ and $v_{k,r}$ for $r=1,...,n$ for  broadcast. 
\item Each user $r$ collects $\widehat x_{k,i}$ from his neighbors $\mathcal{N}_r=\{i:\mathcal{L}_i\cap\mathcal{L}_r\neq\emptyset\}$ and {\small$\nabla^2 B_l\big(\sum_{u:l\in\mathcal{L}_u}\widehat x_{k,u}\big)$} from its links $l\in\mathcal{L}_r$.
\end{itemize}}

               \STATE{For each user $r$, update auxiliary variable $v_{k,r}$ by \cref{v:updates}.
               }
               \vspace{0.1cm}
               \STATE{For each user $r=1,...,n$, update user-specified parameters $ \alpha_{k+1,r}$ by \cref{a:update}.
               }                          
		\ENDFOR
	\end{algorithmic}
	\end{algorithm}

As shown in \cref{alg:bio}, this algorithm involves a two-level optimization procedures. 
For the lower level, a standard distributed primal algorithm (examples can be found in  \cite{srikant2013communication}) is used to get $\delta_\Phi$-approximated solutions $\widehat x_{k,r}$ such that $|\widehat x_{k,r}-x_{k,r}^*|\leq \delta_\Phi$ ($\delta_\Phi$ is sufficiently small) under $\alpha_{k,r}$ for $r=1,...,n$ , where $x_{k,r}^*,r=1,...,n$ are the lower-level solutions of the problem \cref{eq:biobj} and are given by
\begin{align*}
x_{k,1}^*,....,x_{k,n}^* =& \argmax_{x_1,...,x_n>0}\;\; \sum_{r=1}^n\big (U_r(x_r;\alpha_{k,r})-\frac{\epsilon x_r^2}{2}\big)  - \sum_{l\in\mathcal{L}}B_l\Big(\sum_{i:l\in\mathcal{L}_i}x_i\Big).
\end{align*}
Note that the above solutions $\widehat x_{k,r},r=1,...,n$ are achievable even in the presence of network delays as long as the execution time $T_l$ is long enough~\citep{YinSriEry_07}. 

For the next stage, all users continue to transmit packages to broadcast their information $\widehat x_{k,r}$ and $v_{k,r}$ for $r=1,...,n$ over the network. Each user stops broadcast once he receives all information $\widehat x_{k,i}$ from his  neighbors $\mathcal{N}_r=\{i:\mathcal{L}_i\cap\mathcal{L}_r\neq\emptyset\}$ (including himself) and constraint-induced quantities {\small$\nabla^2 B_l\big(\sum_{u:l\in\mathcal{L}_u}\widehat x_{k,u}\big)$} from all links $l\in\mathcal{L}_r$ along his path. This process is finished after a sufficiently long time $T_o$, i.e., no packages are transmitted in the networks. Note that each user can easily distinguish packages in this stage from those in the previous NUM procedure via identifying the existence of the new variable $v_{k,r}$. 


After receiving the neighbor information $\widehat x_{k,i},v_{k,i}$ for $i\in\mathcal{N}_r$, each user $r$ update the auxiliary variable $v_{k,r}$ locally by 
\begin{align}\label{v:updates}
               v_{k+1,r} =& -\eta\sum_{i\in\mathcal{N}_r}\sum_{l\in\mathcal{L}_i\cap\mathcal{L}_r} \nabla^2 B_l\Big(\sum_{j:l\in\mathcal{L}_j}\widehat x_{k,j}\Big)v_{k,i} \nonumber
               \\&+\big(1-\epsilon+\eta\nabla_{x}^2U_r(\widehat x_{k,r};\alpha_{k,r})\big)v_{k,r} - \underbrace{\eta\nabla\widetilde U_r(\widehat x_{k,r})}_{\text{user feedback}},
\end{align}
where the important quantity {\small$\nabla \widetilde U_r(\widehat x_{k,r})$} reflects how fast the user experience can increase when increasing the current supply $\widehat x_{k,r}$. Note that the update in \cref{v:updates} for user $r$ only uses the information of its neighbors with at least one common link, so it is amenable to the practical decentralized implementation. The updates in \cref{v:updates} for $r=1,...,n$ can be regarded as one-step approximation of the Hessian-inverse-vector product {\small$\big( \nabla^2_{\bm{x}} \Phi(\bm{x}_k^*;\bm{\alpha}_k) \big)^{-1}[\nabla \widetilde U_1(x^*_{k,1}),...,\nabla \widetilde U_n(x^*_{k,n})]^T$} of the hypergradient in \cref{eq:hyperGrad}. The quantity {\small$\nabla\widetilde U_r(\widehat x_{k,r})$} of \cref{v:updates} is constructed via querying the use experience on the received resource. As mentioned before, we use the gradient-type information from users to improve the resource allocation via asking how fast their experiences increase when supplying slightly more resource than $\widehat x_{k,r}$. In some circumstances where only utility values are observed, e.g., user satisfaction or job quality,  we also provide a derivative-free gradient approximation   
using only utility values $\widetilde U_r(\cdot)$ in \Cref{zeroht-order_sec}.

 Finally,  each user $r$ updates the user-specified parameter $\alpha_{k,r}$  via a projected gradient ascend step as
\begin{align}\label{a:update}
               \alpha_{k+1,r}=\mathcal{P}_{\mathcal{A}_r}\Big\{\alpha_{k,r}-\beta\nabla_{\alpha}\nabla_{x} U_r(\widehat x_{k,r};\alpha_{k,r})v_{k+1,r}\Big\},
\end{align}
where $\beta>0$ is the outer-loop stepsize and $\mathcal{P}_{\mathcal{A}_r}(\cdot)$ is the projection onto the constraint set $\mathcal{A}_r$.
\subsection{Main Results}
We present the finite-time convergence analysis for our proposed distributed method in \cref{alg:bio}. 
 We first introduce some definitions and assumptions. 
\begin{Definition}
$f(z): \mathcal{Z}\rightarrow \mathbb{R}^d$ is $L$-Lipschitz continuous if for $\forall z_1,z_2\in\mathcal{Z}$, $\|f(z_1)-f(z_2)\|\leq L\|z_1-z_2\|$. 
\end{Definition}
Without loss of generality, We make the following assumptions on the objective function in \cref{eq:biobj}.
\begin{Assumption}\label{assum:boundX}
The lower-level solution $\bm{x}^*(\bm{\alpha})$ in \cref{eq:biobj} is bounded in the sense that  there exist constants $\delta,b>0$ such that its each coordinate satisfies $\delta<x^*_r(\bm{\alpha})<b,\,r=1,...,n$ for $\forall\, \bm{\alpha}\in\mathcal{A}$. 
\end{Assumption}
Assumption~\ref{assum:boundX} says that the lower-level solutions $x_r^*,r=1,...,n$ are lower and upper-bounded by a small constant $\delta>0$ and a sufficiently large constant $b$. This assumption is reasonable because the regularization $B_l(\cdot)$ prevents the solutions from converging to the infinity and the lower bound constant $\delta$ helps to avoid some trouble when $x_r\rightarrow 0$ for some utility function such as $\log(x_r)$ and $\frac{x_r^{1-\alpha_r}}{1-\alpha_r}$ with $\alpha
_r>1$. 
For example, it can be shown that the solutions of for $\alpha$-fairness utility function $U_r(x_r;\alpha_r)=\frac{x_r^{1-\alpha_r}}{1-\alpha_r}$ satisfies Assumption~\ref{assum:boundX} given the boundedness of $\bm{\alpha}\in\mathcal{A}$.

The following assumption imposes some geometrical conditions on the utility function $U_r(\cdot\,;\alpha_r)$ and the regularization function  $B_l(\cdot)$. Let $\mathcal{X}:=\{\bm{x}:\frac{\delta}{2}<x_r<2b,r=1,...,n\}$.
\begin{Assumption}\label{assum:geometry}
For any $\bm{\alpha}\in\mathcal{A}$  and any $\bm{x}\in\mathcal{X}$,
\begin{list}{$\bullet$}{\topsep=0.1ex \leftmargin=0.15in \rightmargin=0.1in \itemsep =0.04in}
\item $U_r(\cdot\,;\alpha_r)$ is concave and $B_l(\cdot)$ is $\mu$-strongly-convex. 
\item $\widetilde U_r(\cdot)$, $\nabla\widetilde U_r(\cdot)$, $\nabla_{x}U_r(\cdot\,;\cdot)$, $\nabla_\alpha\nabla_x U_r(\cdot\,;\cdot)$,  $\nabla_x^2 U_r(\cdot\,;\cdot)$ are $L_u$-Lipschitz continuous. 
\item $\nabla B_l(\cdot)$ and $\nabla^2B_l(\cdot)$ are $L_b$-Lipschitz continuous.
\end{list}
\end{Assumption}
Assumption~\ref{assum:geometry} cover many utility functions of practical interest such as log utility $\log(x_r)$ and $\alpha$-fairness utility, as well as a variety of regularizers such as the quadratic function $\frac{\mu}{2}x^2$ and the barrier function $-\log(c-x)$ for $\frac{\delta}{2}<x< 2b<c$. For example, for the 
$\alpha$-fairness utility function  $\frac{x_r^{1-\alpha_r}}{1-\alpha_r}$, the Lipschitz continuity assumption holds because its high-order derivatives such as  $\nabla_x U_r(x_r;\alpha_r)=\frac{1}{x_r^{\alpha_r}}$, $\nabla_x^2U_r(x_r;\alpha_r)=\frac{-\alpha_r}{x_r^{\alpha_r+1}}$, $\nabla_x^3U_r(x_r;\alpha_r)= \frac{\alpha_r(\alpha_r+1)}{x_r^{\alpha_r+2}}$ are {\bf bounded} due to the boundedness of $\alpha_r\in\mathcal{A}_r$ and $x_r\in(\frac{\delta}{2},2b)$.  

The following theorem characterizes the convergence rate analysis for the proposed algorithm with general utility functions and networks.
\begin{theorem}\label{th:mainConvergence}
Suppose Assumptions~\ref{assum:boundX} and~\ref{assum:geometry} hold. Choose  $\delta_\phi < \frac{\delta}{2}$, $\eta<\frac{1}{L_{\text{grad}}}$ and $\beta\leq\min\big(\sqrt{\frac{\eta\mu_\Phi}{256C_vL_u^2}},\frac{1}{2L_\Psi}\big)$, where $L_\Psi=\big(\frac{L_{\text{grad}}}{\mu_\Phi}
 \big( \frac{\sqrt{n}L_u^2}{\mu_\Phi} + \frac{\sqrt{n}L_u^2L_{\text{Hess}}}{\mu_\Phi^2}+\frac{L_u}{\mu_\Phi}\big) + \frac{\sqrt{n}L_u^2}{\mu_\Phi} + \frac{\sqrt{n}L_u^3}{\mu_\Phi^2} \big)$ is the smoothness constant of the total objective function $\Psi(\bm{\alpha})$. Then, the iterates generated by \Cref{alg:bio} satisfy 
\begin{align*}
\frac{1}{K}\sum_{k=0}^{K-1}\|G_{\text{proj}}(\bm{\alpha}_k)\|^2\leq & \underbrace{\frac{16(\max_{\bm{\alpha}\in\mathcal{A}}\Psi(\bm{\alpha})-\Psi(\bm{\alpha}_0))}{\beta K} +\frac{256nL_u^4(1+\mu_\Phi^{2})}{\eta\mu_\Phi^3} \frac{1}{K} }_{\text{Sublinearly decaying terms}}
\\&+\underbrace{\frac{128L_u^2C_\Phi\delta^2_\Phi}{\eta\mu_\Phi}+\frac{4 L^4_u n^2\delta^2_\Phi}{\mu^2_\Phi}}_{\text{Lower-level error}},
\end{align*}
where $G_{\text{proj}}(\bm{\alpha}_k)=\beta^{-1}( \mathcal{P}_{\mathcal{A}}\{ \bm{\alpha}_{k} +\beta \nabla \Psi(\bm{\alpha}_{k})\}-\bm{\alpha}_{k})$ denote the generalized projected gradient at the $k^{th}$ iteration, and 
$\mu_\Phi$, $L_{\text{grad}}$, $L_{\text{Hess}}$, 
$C_\Phi$ and $C_v$ are the constants defined in Propositions \ref{prop:sc}, \ref{prop:lipschitz_Phi} and \ref{prop:v_updates}, respectively.
\end{theorem}
 \Cref{th:mainConvergence} uses the generalized gradient $G_{\text{proj}}(\bm{\alpha}_k)$ instead of the gradient $\nabla \Psi(\bm{\alpha}_{k})$ due to the existence of the projection. Note that if the iterate $ \bm{\alpha}_{k} 
+\beta \nabla \Psi(\bm{\alpha}_{k})$ locates inside of the constraint set $\mathcal{A}$, this generalized gradient  $G_{\text{proj}}(\bm{\alpha}_k)$ reduces to the vanilla gradient $ \nabla \Psi(\bm{\alpha}_k)$. 

\Cref{th:mainConvergence} shows that the proposed DBiNUM finds a stationary point $\bm{\alpha}_{s}$ with $s=\argmin_{k}\|G_{\text{proj}}(\bm{\alpha}_k)\|^2$
for the constrained nonconcave bilevel problem in \cref{eq:biobj}, whose generalized projected gradient norm  $\|G_{\text{proj}}(\bm{\alpha}_s)\|^2$ 
contains a sublinearly decaying term   and  
a convergence error $\frac{128L_u^2C_\Phi\delta_\Phi^2}{\eta\mu_\Phi}+\frac{4 L^4_u n^2\delta^2_\Phi}{\mu^2_\Phi}$ induced by the approximation error $\delta_\Phi$  of the lower-level network utility maximization. This convergence error can be arbitrarily small by setting the lower-level target accuracy $\delta_\Phi$  small, e.g., at an $\epsilon$ accuracy. Note that we adopt the stationary point as the convergence criterion due to the general nonconcavity  of the upper-level objective function $\widetilde U_r$. 

\section{Proof of the Main Result}
In this section, we provide the technical proofs for \Cref{th:mainConvergence}. 
We first prove an important strongly-concave geometry of the lower-level objective function $\Phi(\mathbf{x};\bm{\alpha})$.
\begin{proposition}\label{prop:sc}
Suppose Assumptions~\ref{assum:geometry} holds. For  any $\bm{\alpha}\in\mathcal{A}, \bm{x}\in\mathcal{X}$,  $\Phi(\bm{x}\,;\bm{\alpha})$ is $\mu_{\Phi}$-strongly-concave w.r.t.~$\bm{x}$, where $\mu_{\Phi}:=\frac{\epsilon+\mu M_{\min}}{2}$ with $M_{\min}=\min_{r=1,...,n}\{M_r: \text{number of links user $r$ exclusively occupies}\}$.\end{proposition}
Note that the strong-concavity constant $\frac{\epsilon+\mu M_{\min}}{2}$ depends on the network topology due to the factor $M_{\min}$. For the case where each user $r$ occupies solely at least one link, $M_{\min}\geq 1$ and hence the quadratic term $\frac{\epsilon}{2}x_r^2,r=1,...,n$ in \cref{eq:biobj} are not needed. However, for the general topology, this quadratic regularization is necessary to guarantee the strong-concavity.  

In the worst cases, the smoothness parameter of the Hessian matrix $\nabla^2_{\bm{x}} \Phi(\cdot\,;\bm{\alpha})$ whose form is given by \Cref{prop:hyperG} scales in the order of $n^{\frac{3}{2}}|\mathcal{L}|$, which can be prohibitively large in the network with millions of users and links, and hence leads to slow convergence in practice. For this reason, we next provide a refined analysis of the smoothness of quantities $\nabla^2_{\bm{x}} \Phi(\bm{x};\bm{\alpha})$ and $ \nabla_{\bm{\alpha}} \nabla_{\bm{x}} \Phi(\bm{x};\bm{\alpha})$ by taking the sparse network structure (i.e., each user shares links with only some of other users) into account. 
\begin{proposition}\label{prop:lipschitz_Phi} 
Suppose Assumption~\ref{assum:geometry} holds. Then, for any $\bm{\alpha}\in\mathcal{A}$, $\bm{x}\in\mathcal{X}$ and any vector $\bm{u}=[u_1,...,u_n]$,
{\small
\begin{align*}
\|\nabla_{\bm{x}} \Phi(\bm{x};\bm{\alpha})-\nabla_{\bm{x}} \Phi(\bm{x}^\prime;\bm{\alpha})\| \leq & L_{\text{grad}}\|\bm{x}-\bm{x}^\prime\|, \nonumber
\\\|\nabla^2_{\bm{x}} \Phi(\bm{x};\bm{\alpha})\bm{u} - \nabla^2_{\bm{x}} \Phi(\bm{x}^{\prime};\bm{\alpha})\bm{u}\|\leq & L_{\text{Hess}}\max_{i}|u_i| \|\bm{x}-\bm{x}^\prime\|,
\nonumber
\\\|\nabla^2_{\bm{x}} \Phi(\bm{x};\bm{\alpha})\bm{u} - \nabla^2_{\bm{x}} \Phi(\bm{x};\bm{\alpha}^\prime)\bm{u}\| \leq &L_u\max_{i}|u_i| \|\bm{\alpha}-\bm{\alpha}^\prime\|, 
\end{align*}}
\hspace{-0.12cm}where {\small $L_{\text{grad}}=\sqrt{2L_u^2 +2n\sum_{i=1}^n\sum_{l:l\in\mathcal{L}_i}L_b^2 } $} and {\small$L_{\text{Hess}}:=\sqrt{ 2L_u^2 + 2n L_b^2\max_i\Big(\sum_{j:\mathcal{L}_i\cap\mathcal{L}_j\neq\emptyset}\sum_{l\in\mathcal{L}_i\cap\mathcal{L}_j} 1  \Big)^2 }$} are constants related to the network topology. Similarly, for the mixed derivative $ \nabla_{\bm{\alpha}} \nabla_{\bm{x}} \Phi(\bm{x};\bm{\alpha})$, we have 
{\small
\begin{align*}
\|\nabla_{\bm{\alpha}} \nabla_{\bm{x}} \Phi(\bm{x};\bm{\alpha}) \bm{u}-\nabla_{\bm{\alpha}} &\nabla_{\bm{x}} \Phi(\bm{x}^\prime;\bm{\alpha})\bm{u} \|\leq L_u\max_{i}|u_i^2|\|\bm{x}-\bm{x}^\prime\|, \nonumber
\\\|\nabla_{\bm{\alpha}} \nabla_{\bm{x}} \Phi(\bm{x};\bm{\alpha}) \bm{u}-\nabla_{\bm{\alpha}} &\nabla_{\bm{x}} \Phi(\bm{x};\bm{\alpha}^{\prime})\bm{u} \|\leq  L_u\max_{i}|u_i|\|\bm{\alpha}-\bm{\alpha}^\prime\|.
\end{align*}}
\end{proposition}
It can be observed from \Cref{prop:lipschitz_Phi}  that the smoothness constant of $\nabla^2_{\bm{x}} \Phi(\cdot\,;\bm{\alpha})\bm{u}$
 scales in the order of $\sqrt{n\max_i (\sum_{j:\mathcal{L}_i\cap\mathcal{L}_j\neq\emptyset}\sum_{l\in\mathcal{L}_i\cap\mathcal{L}_j} 1)^2}$, which represents to the total number of links the users $i$ share with other users. As mentioned before, In the worst case, i.e., all users share the same links, this constant takes the order of $n^{\frac{3}{2}}|\mathcal{L}|$. However, in the practical network, each user shares links with a small portion of users, and hence each $\sum_{j:\mathcal{L}_i\cap\mathcal{L}_j\neq\emptyset}\sum_{l\in\mathcal{L}_i\cap\mathcal{L}_j} 1$ is much smaller than the worst-case $n|\mathcal{L}|$. 
 
 We next characterize the error in approximating the Hessian-inverse-vector product in the hypergradient at  iteration $k$. For notational convenience, let {\small$\bm{v}_k=[v_{k,1},...,v_{k,n}]^T$} and {\small$\nabla\widetilde U(\bm{x})= \big[\nabla \widetilde U_1(x_1),...,\nabla \widetilde U_n(x_n)\big]^T$}.  

 \begin{proposition}\label{prop:v_updates}
 Suppose Assumptions~\ref{assum:boundX} and~\ref{assum:geometry} hold. Choose  $\delta_\phi < \frac{\delta}{2}$ and $\eta<\frac{1}{L_{\text{grad}}}$. Let $C_v=n\big(1+\frac{2}{\eta\mu_\Phi}\big) \big( \frac{L_{\text{grad}}}{\mu_\Phi}\big(\frac{ L_uL_{\text{Hess}}}{\mu_\Phi^2} +\frac{L_u}{\sqrt{n}\mu_\Phi} \big) + \frac{L_u^2}{\mu_\Phi^2} \big) ^2$ and $C_\Phi=4\big(1+\frac{1}{\eta\mu_\Phi} \big)\big(\frac{L_{\text{Hess}}L_u}{\mu_\Phi^2} +\frac{L_u}{\sqrt{n}\mu_\Phi} \big)^2 n^2$. Then, we have
\begin{align}\label{eq:iter_v_re}
 &\| \bm{v}_{k+1} - ( \nabla^2_{\bm{x}} \Phi(\bm{x}^*_k;\bm{\alpha}_k) )^{-1}\nabla\widetilde U(\bm{x}_k^*) \|^2  \nonumber
 \\&\leq \big(1-\frac{\eta\mu_\Phi}{2}\big)\big\|  \bm{v}_{k} - \nabla^2_{\bm{x}} \Phi(\bm{x}^*_{k-1};\bm{\alpha}_{k-1}) ^{-1}\nabla\widetilde U(\bm{x}_{k-1}^*)  \big\|^2 \nonumber
 \\&+ C_v\|\bm{\alpha}_k-\bm{\alpha}_{k-1}\|^2 +C_\Phi\delta_\Phi^2.
\end{align}  
 \end{proposition}
\Cref{prop:v_updates} characterizes the error of $\bm{v}_{k+1}$ in approximating the Hessian-inverse-vector product  {\small$( \nabla^2_{\bm{x}} \Phi(\bm{x}^*_k;\bm{\alpha}_k) )^{-1}\nabla\widetilde U(\bm{x}_k^*)$} of the hypergradient. It can be seen from \cref{eq:iter_v_re} that this error contains an iteratively decreasing term (i.e., the first term at the right hand side) and two error terms $C_v\|\bm{\alpha}_k-\bm{\alpha}_{k-1}\|$ (which captures the difference between two adjacent iterations) and $C_\Phi\delta_\Phi^2$ (which is induced by the lower-level estimation error $\|\bm{\widehat x}_k-\bm{x}_k\|$). By choosing the upper-level stepsize $\beta$ small enough, we can well control the increment $\|\bm{\alpha}_k-\bm{\alpha}_{k-1}\|$ and guarantee the hypergradient estimation error not to explode.  Based on the form of the hypergradient established in \Cref{prop:hyperG}, the update in \cref{a:update} can be written as 
\begin{align}\label{eq:alpha_updates}
\bm{\alpha}_{k+1} = \mathcal{P}_{\mathcal{A}}\big\{ \bm{\alpha}_{k} -\beta \nabla_{\bm{\alpha}} \nabla_{\bm{x}} \Phi(\bm{\widehat x}_k;\bm{\alpha}_k) \bm{v}_{k+1}   \big\},
\end{align}
where $\widehat\nabla \Psi(\bm{\alpha}_k) :=-\nabla_{\bm{\alpha}} \nabla_{\bm{x}} \Phi(\bm{\widehat x}_k;\bm{\alpha}_k) \bm{v}_{k+1} $ serves as an estimator of the hypergradient $\nabla \Psi(\bm{\alpha}_k) $ given by \cref{eq:hyperGrad}. 
 We now characterize the error between $\widehat\nabla \Psi(\bm{\alpha}_k)$ and $\nabla \Psi(\bm{\alpha}_k) $. 
  \begin{proposition}\label{prop:hyperGapproximate}
  Suppose Assumptions~\ref{assum:boundX} and~\ref{assum:geometry} hold. Choose  $\delta_\phi < \frac{\delta}{2},\eta<\frac{1}{L_{\text{grad}}}$ and $\beta<\sqrt{\frac{\eta\mu_\Phi}{16C_vL^2_u}}$. Then,
  \begin{align}\label{eq:hyperBoundForm}
  \big\|\widehat\nabla \Psi(\bm{\alpha}_k)- \nabla \Psi(\bm{\alpha}_k) \big\|^2  \leq &\big(1-\frac{\eta\mu_\Phi}{4}\big)^{k} 4nL_u^4(1+\mu_\Phi^{-2})  \nonumber
  \\& +4C_vL_u^2\beta^2\sum_{t=0}^{k-1}\big(1-\frac{\eta\mu_\Phi}{4}\big)^{k-1-t}\|G_{\text{proj}}(\bm{\alpha}_{t})\|^2 \nonumber
\\& +\frac{8L_u^2C_\Phi\delta_\Phi^2\mu_\Phi+  4\eta L^4_u n^2\delta^2_\Phi}{\eta\mu^2_\Phi}, 
  \end{align}
  where the constants $C_v,C_\Phi$ are given in \Cref{prop:v_updates}, 
 \end{proposition}
\Cref{prop:hyperGapproximate} shows that the bound on the hypergradient estimation error $ \big\|\widehat\nabla \Psi(\bm{\alpha}_k)- \nabla \Psi(\bm{\alpha}_k) \big\|^2$ contains three terms, i.e., an exponentially decaying term, an error term proportional to the average gradient norm, and   
a sufficiently small error term induced by the lower-level approximation. 
Based on the results in Propositions~\ref{prop:sc},  \ref{prop:lipschitz_Phi},\ref{prop:v_updates} and \ref{prop:hyperGapproximate}, we now characterize the convergence rate performance of the distributed bilevel method in \Cref{alg:bio}. 

\begin{proof}[{\bf Proof Sketch of \Cref{th:mainConvergence}}] The first step is to derive the smoothness property of the hypergradient $ \nabla \Psi(\cdot)$. Based on the form of $ \nabla \Psi(\cdot)$ in \cref{eq:hyperGrad} and using  \Cref{prop:sc}, \Cref{prop:lipschitz_Phi}, we have, for any two $\bm{\alpha}_1,\bm{\alpha}_2\in\mathcal{A}$
\begin{align}\label{eq:smoothPhi}
\|\nabla \Psi(\bm{\alpha}_1) - \nabla \Psi(\bm{\alpha}_2)\| \leq & L_u(\|\bm{x}^*(\bm{\alpha}_1)-\bm{x}^*(\bm{\alpha}_2)\|+\|\bm{\alpha}_1-\bm{\alpha}_2\|) \frac{\sqrt{n}L_u}{\mu_\Phi}  \nonumber
\\&+ \frac{\sqrt{n}L^2_u}{\mu_\Phi^2}\big(L_{\text{Hess}}\|\bm{x}^*(\bm{\alpha}_1)-\bm{x}^*(\bm{\alpha}_2)\|+L_u \|\bm{\alpha}_1-\bm{\alpha}_2\|\big) \nonumber
\\&+ \frac{L_u}{\mu_\Phi}\|\bm{x}^*(\bm{\alpha}_1)-\bm{x}^*(\bm{\alpha}_2)\|,
\end{align} 
which, using $\|\bm{x}^*(\bm{\alpha}_1)-\bm{x}^*(\bm{\alpha}_2)\|\leq \frac{L_{\text{grad}}}{\mu_\Phi}\|\bm{\alpha}_1-\bm{\alpha}_{2}\|$, 
yields
\begin{align}\label{eq:smoothnPsiF}
\|\nabla \Psi(\bm{\alpha}_1) - \nabla \Psi(\bm{\alpha}_2)\|\leq L_\Psi\|\bm{\alpha}_1-\bm{\alpha}_2\|.
\end{align}
Let  {\small$\widehat \nabla \Psi(\bm{\alpha}_k)=-\nabla_{\bm{\alpha}}\nabla_{\bm{x}} \Phi(\bm{\widehat x}_k;\bm{\alpha}_k) \bm{v}_{k+1}$} demote the hypergradient estimate. Then,  
based on the smoothness property established in \cref{eq:smoothnPsiF}, we have 
\begin{align}\label{eq:smoothness_psi}
\Psi(\bm{\alpha}_{k+1})-\Psi(\bm{\alpha}_{k}) &\geq    \Big\langle \widehat \nabla \Psi(\bm{\alpha}_k),  \mathcal{P}_{\mathcal{A}}\big\{\bm{\alpha}_k+\beta \widehat \nabla \Psi(\bm{\alpha}_k) \big\} -\bm{\alpha}_k\Big\rangle  \nonumber
\\&\quad+ \Big\langle\nabla \Psi(\bm{\alpha}_k)-\widehat \nabla \Psi(\bm{\alpha}_k),\mathcal{P}_{\mathcal{A}}\big\{\bm{\alpha}_k+\beta \widehat \nabla \Psi(\bm{\alpha}_k) \big\} -\bm{\alpha}_k \Big\rangle \nonumber
\\&\quad- \frac{L_\Psi}{2}\|\bm{\alpha}_{k+1}-\bm{\alpha}_{k}\|^2. 
\end{align}
Using the property of the projection on the convex set $\mathcal{A}$, i.e., {\small$\langle\bm{x}-\mathcal{P}_{\mathcal{A}}(\bm{x}),\bm{y}-\mathcal{P}_{\mathcal{A}}(\bm{x})\rangle\leq 0$} for any $\bm{y}\in\mathcal{A}$ and noting that $\bm{\alpha}_k\in\mathcal{A}$, the first term of the right hand side of \cref{eq:smoothness_psi} can be lower-bounded by
\begin{align}\label{eq:posiPart}
\frac{1}{\beta} \|\bm{\alpha}_k- \mathcal{P}_{\mathcal{A}}\big\{\bm{\alpha}_k+\beta \widehat \nabla \Psi(\bm{\alpha}_k) \big\}\|^2.
\end{align}
Let $\widehat G_{\text{proj}}(\bm{\alpha}_k)=\beta^{-1}\big(\mathcal{P}_{\mathcal{A}}\big\{\bm{\alpha}_k+\beta \widehat \nabla \Psi(\bm{\alpha}_k) \big\}-\bm{\alpha}_k\big)$ be the estimate of the generalized projected gradient {\small$G_{\text{proj}}(\bm{\alpha}_k)$} defined in \Cref{prop:hyperGapproximate}. Then,  
substituting \cref{eq:posiPart} into \cref{eq:smoothness_psi} and based on $\langle \bm{a}, \bm{b}\rangle \geq -\frac{1}{2}(\|\bm{a}\|^2 + \|\bm{b}\|^2)$ and the non-expansive property of the projection on convex sets, we have  
\begin{align}\label{eq:smoothInterM}
\Psi(\bm{\alpha}_{k+1}) \geq & \Psi(\bm{\alpha}_{k}) + \big(\frac{\beta}{4} -\frac{L_\Psi\beta^2}{4}\big)\|G_{\text{proj}}(\bm{\alpha}_k)\|^2 \nonumber
\\&-  \big(\beta -\frac{L_\Psi\beta^2}{2}\big) \|\nabla \Psi(\bm{\alpha}_k)-\widehat \nabla \Psi(\bm{\alpha}_k)\|^2.
 \end{align}
 Applying \Cref{prop:hyperGapproximate} to the above \cref{eq:smoothInterM}, conducting the telescoping and using the fact that {\small$\sum_{k=1}^{K-1}\sum_{t=0}^{k-1}a_{k-1-t}b_t\leq \sum_{k=0}^{K-1}a_k\sum_{t=0}^{K-1}b_t$ for $a_t,b_t\geq 0$}, we have  
\begin{align*}
 \Big(\frac{1}{8}  - \frac{16C_vL_u^2\beta^2}{\eta\mu_\Phi}\Big)\frac{1}{K}\sum_{k=0}^{K-1}\|G_{\text{proj}}(\bm{\alpha}_k)\|^2 \leq &\frac{\max_{\bm{\alpha}\in\mathcal{A}}\Psi(\bm{\alpha})-\Psi(\bm{\alpha}_0)}{\beta K} \nonumber
\\ &+\frac{16nL_u^4(1+\mu_\Phi^{-2})}{\eta\mu_\Phi} \frac{1}{K} +\frac{8L_u^2C_\Phi\delta_\Phi^2\mu_\Phi+  4\eta L^4_u n^2\delta^2_\Phi}{\eta\mu^2_\Phi},
\end{align*}
which, in conjunction with the choice of $\beta\leq\sqrt{\frac{\eta\mu_\Phi}{256C_vL_u^2}}$, completes the proof. 
\end{proof}

\section{Validation Study of Bilevel Formulation}\label{se:bi2single}
\begin{figure}[!h]
\centering
\includegraphics[width=9cm]{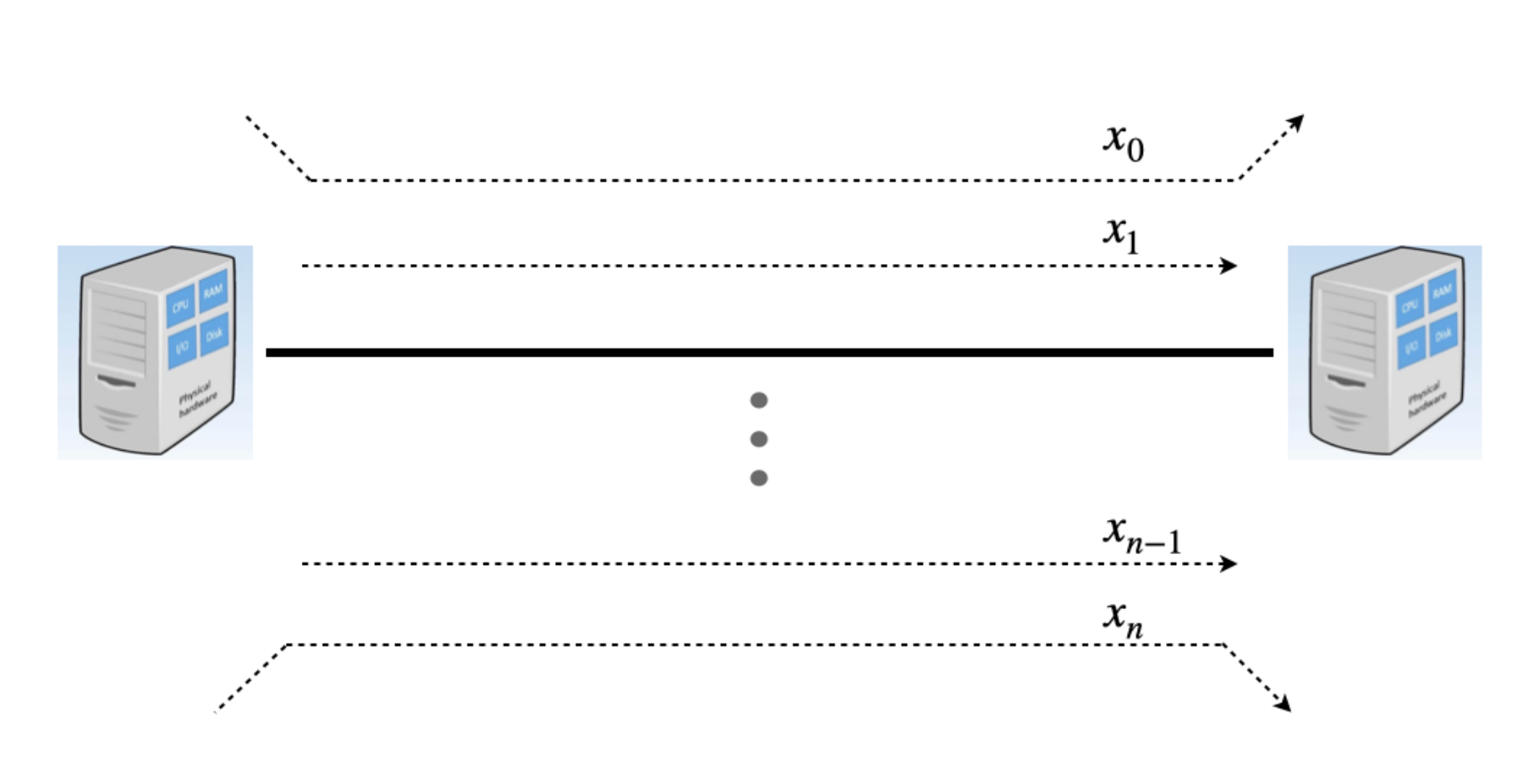}
\caption{Example study: single communication link with $n$ users.}
\label{fig:single_link}
  \end{figure}

In this section, we provide a case study for a single-link multi-user network as shown in \Cref{fig:single_link} to validate the bilevel formulation we propose in \cref{eq:biobj}, where all $n$ users share the same communication link with a capacity $P$. In this setting, the bilevel formulation is solve the following problem. 
\begin{align}\label{bilevel:case_form}
&\max_{\bm{\alpha}\in\mathcal{A}}\;\; \Psi(\bm{\alpha})=\sum_{r=1}^n\frac{x_r^*(\bm{\alpha})^{1-\widetilde \alpha_r}}{1-\widetilde \alpha_r}, \nonumber
\\ & x_1^*(\bm{\alpha}),...,x_n^*(\bm{\alpha}) = \argmax_{x_r > 0, \sum_{r=1}^n x_r \leq P}\;\;\sum_{r=1}^n\frac{x_r^{1- \alpha_r}}{1- \alpha_r},
\end{align}
where we adopt a simple bounded constraint set $\mathcal{A}:=\{0<a_r\leq b, r=1,...,n\}$. 
Note that the lower level adopts the original problem in \cref{eq:obj} rather than the primal version as in \cref{eq:biobj} because the explicit solutions can be obtained here.




The following theorem establishes the  equivalence between the solutions of the bilevel problem in \cref{bilevel:case_form} and the standard single-level NUM,  
when the true user utilities are $\alpha$-fairness functions.
\begin{theorem}\label{th:case_study}
Let $\bm{\alpha}^*\in\argmax_{\bm{\alpha}\in\mathcal{A}}\Psi(\bm{\alpha})$ be any solution of the bilevel problem in \cref{bilevel:case_form}. Then, the resulting allocated resources $x_r^*(\bm{\alpha}^*),r=1,...,n$ from the bilevel formulation recover the solutions of the following standard utility maximization problem under $\alpha$-fairness utilities with fixed parameters $\widetilde\alpha_r>0,r=1,...,n$.
\begin{align}\label{example:test_ori}
\max_{x_1,...,x_n}\;\; \sum_{r=1}^n \Big [\widetilde U_r(x_r;\widetilde \alpha_r)=\frac{x_r^{1-\widetilde \alpha_r}}{1-\widetilde \alpha_r}\Big ],
\end{align} 
subject to 
$\sum_{r=1}^n x_r \leq P\; \text{and} \;x_r > 0, \text{ for } r=1,...,n$. 
\end{theorem}
\Cref{th:case_study} shows that the solution $\bm{x}^*(\bm{\alpha}^*)$ of the bilevel problem we formulate in \cref{bilevel:case_form} also maximizes the original network utility maximization problem in \cref{example:test_ori}. This means that the proposed DBiNUM converges to a solution as if the utility functions are known.  
This case study provides some validation of the proposed bilevel objective function. We note that our analysis is possibly extended to the multi-link scenarios with the graph structure satisfying certain properties.  


\section{Discussion on User Feedback}\label{zeroht-order_sec}
It can be seen from \cref{v:updates} that DBiNUM takes the user (or application) information $\nabla\widetilde U_r(\widehat x_{k,r})$ to improve the selection of the user utility functions. In other words, each user has to give feedback to the network showing how fast their experiences increase at the given allocated resource $\widehat x_{k,r}$. However, in some circumstances, only utility values are available such as energy consumption, user satisfaction or job quality, and hence a more feasible solution is to query their utility value at $x$, i.e.,  $\widetilde U_r(x)$. Given such value information, one can use a gradient-free approach to approximate the gradient $\nabla\widetilde U_r(\widehat x_{k,r})$ by taking the utility difference at two close points $\widehat x_{k,r}$ and $\widehat x_{k,r}+\delta u$, as shown below. 
\begin{align}\label{two_point_estimator}
\widehat \nabla_{\text{two}}\widetilde U_r(\widehat x_{k,r};u) = \frac{\widetilde U_r(\widehat x_{k,r}+\delta u)-\widetilde U_r(\widehat x_{k,r})}{\delta} u,
\end{align}  
where $\delta>0$ is the smoothing parameter and $u$ is a standard Gaussian random variable. Based on the results in \cite{nesterov2017random}, it can be shown that the estimation bias $\big|\mathbb{E}_u\widehat \nabla_{\text{two}}\widetilde U_r(\widehat x_{k,r};u)  -\nabla\widetilde U_r(\widehat x_{k,r}) \big|$of the above two-point estimator is bounded by  $4L_u\delta$, which can be small by choosing a small $\delta$. Hence, we can establish a convergence rate result similar to \Cref{th:mainConvergence} with an error proportional to $\delta$. 

Note that the estimator in \cref{two_point_estimator} requires to query the utility value $\widetilde U_r(\cdot)$ at two points simultaneously. However, in the time-varying and non-stationary environments, $\widetilde U_r$ is changing with time, and hence the two-point estimator may contain large estimation error. In this case, one-point approach turns out to be more appealing, which takes the form of  
\begin{align}
\widehat \nabla_{\text{one}}\widetilde U_r(\widehat x_{k,r};u) = \frac{\widetilde U_r(\widehat x_{k,r}+\delta u)u}{\delta}.
\end{align} 
It can be shown the above one-query estimator has the same mean as the two-query estiamtion, i.e., $\mathbb{E}_u \widehat \nabla_{\text{one}}\widetilde U_r(\widehat x_{k,r};u) = \mathbb{E}_u \widehat \nabla_{\text{two}}\widetilde U_r(\widehat x_{k,r};u) $, so the convergence analysis in \Cref{th:mainConvergence} is still applied. 

\begin{figure*}[ht]
\centering
\includegraphics[width=5cm]{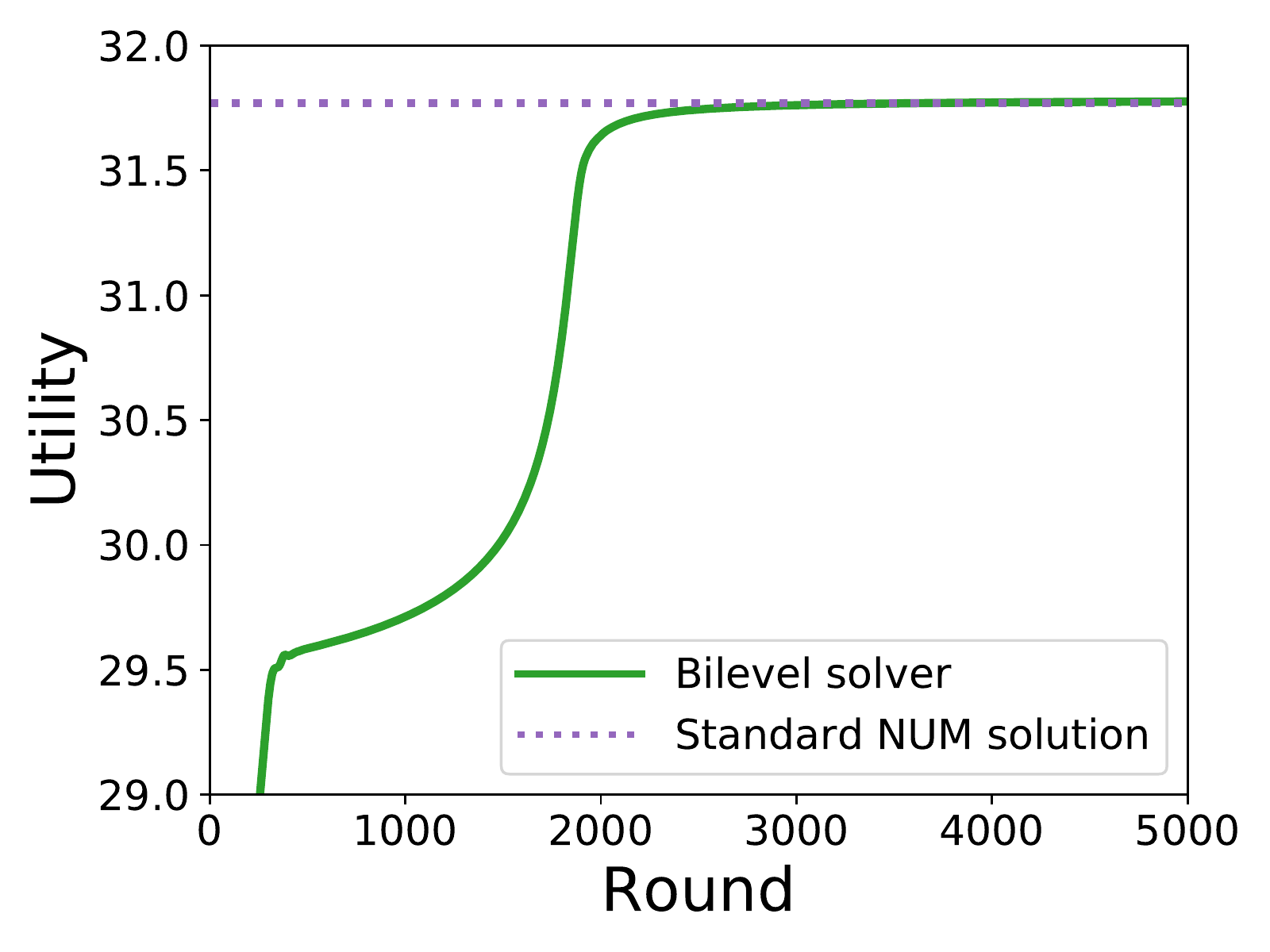}
\includegraphics[width=5cm]{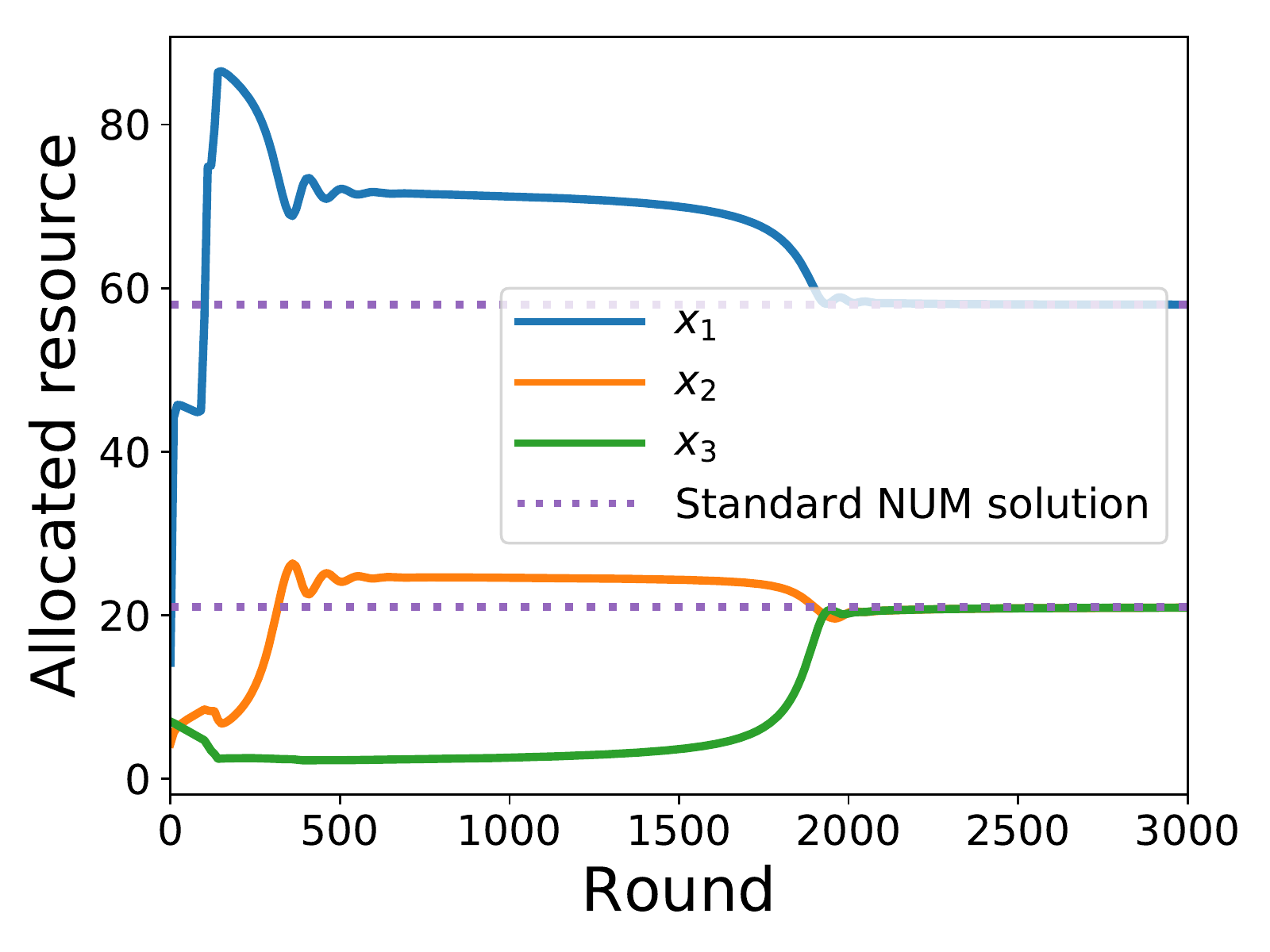}
\includegraphics[width=5cm]{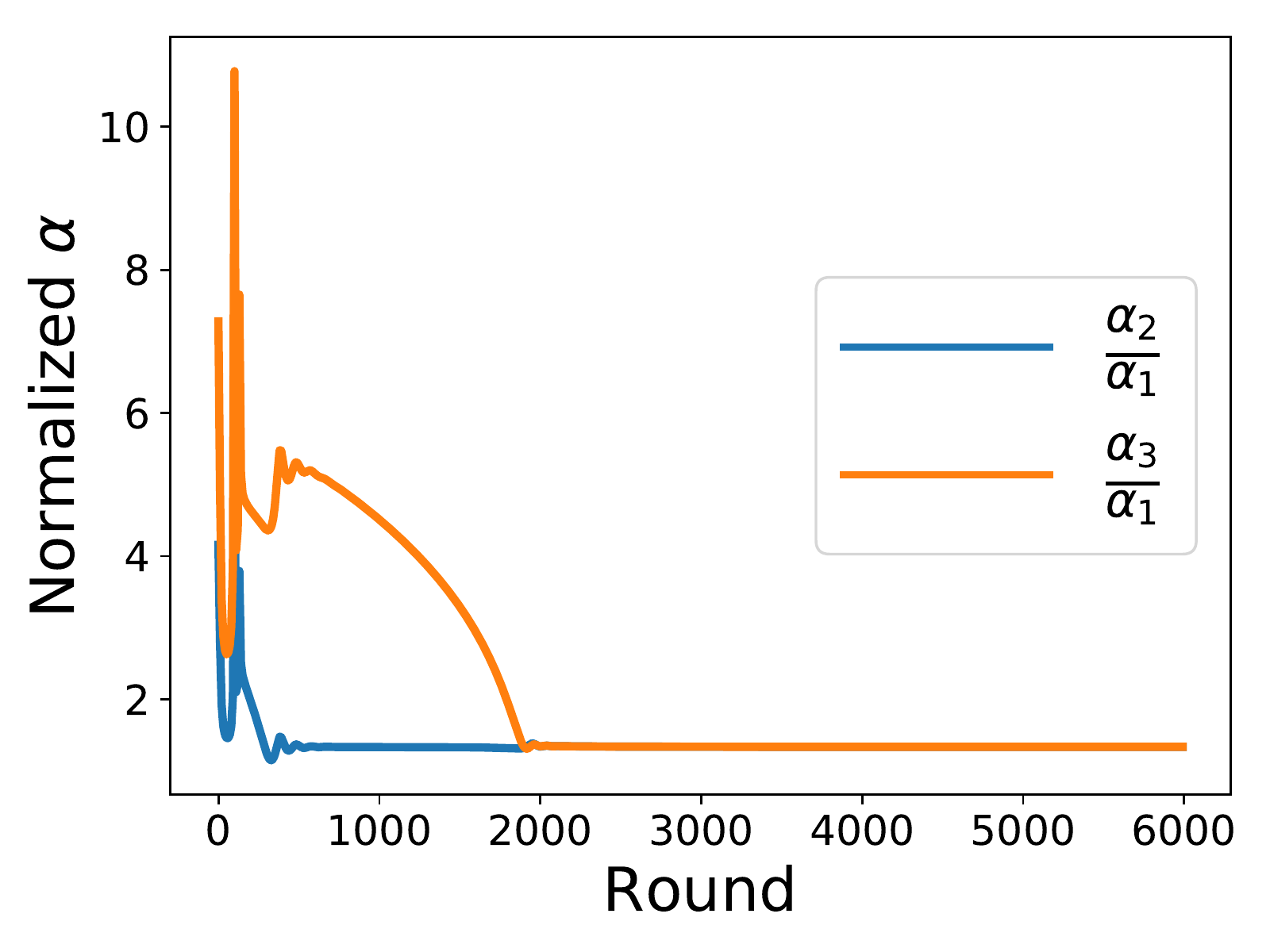}
\caption{Network utility maximization via our proposed bilevel solver DBiNUM in a $3$-user setting. Left plot: total underlying utility $\Psi$ v.s.~\# of rounds; middle plot: allocated resource v.s.~\# of rounds; right plot: normalized $\alpha$ v.s.~\# of rounds.}
\label{exp:simu_1}
  \end{figure*}
  
  \begin{figure*}[ht]
\centering
\includegraphics[width=5cm]{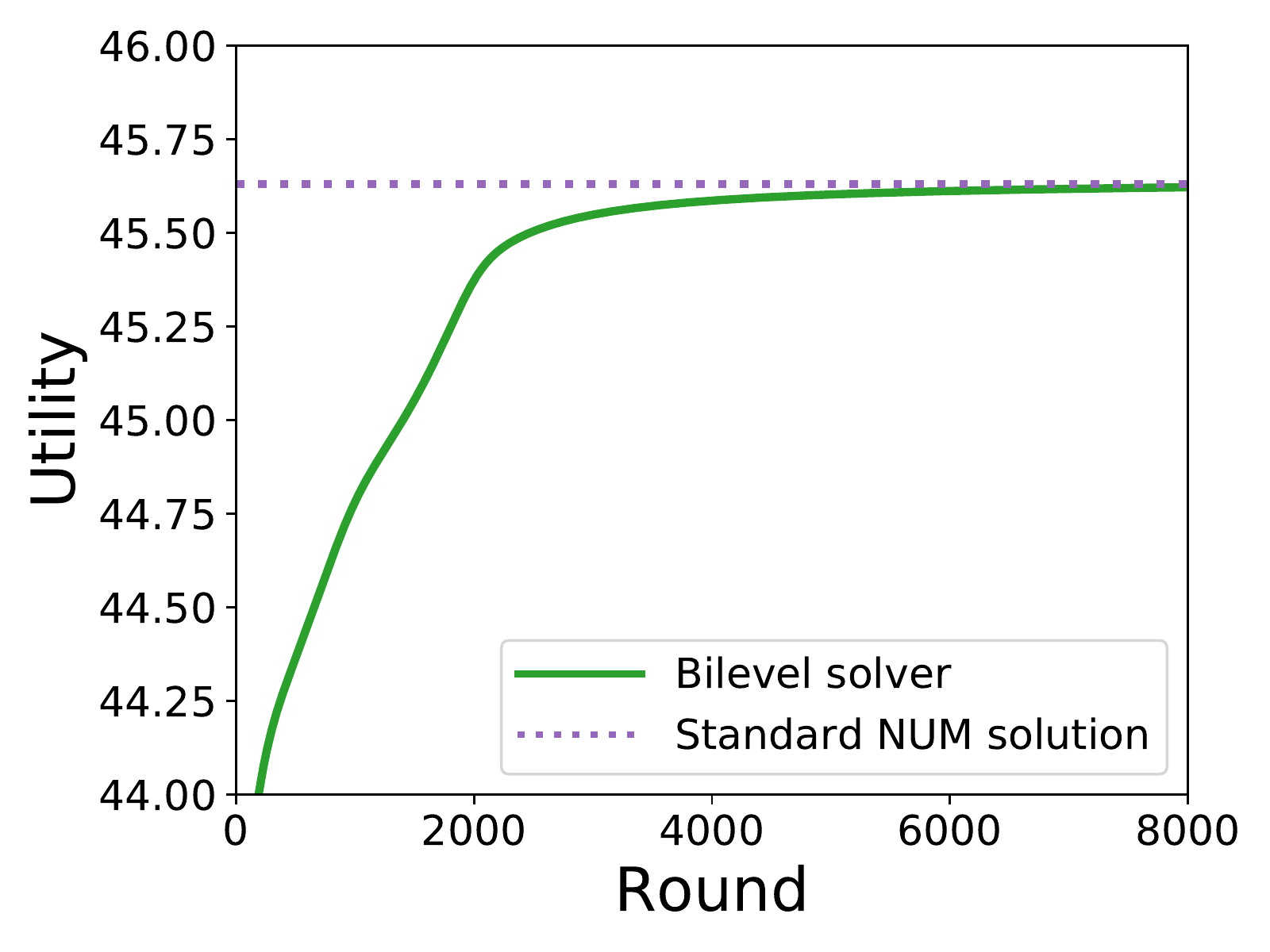}
\includegraphics[width=5cm]{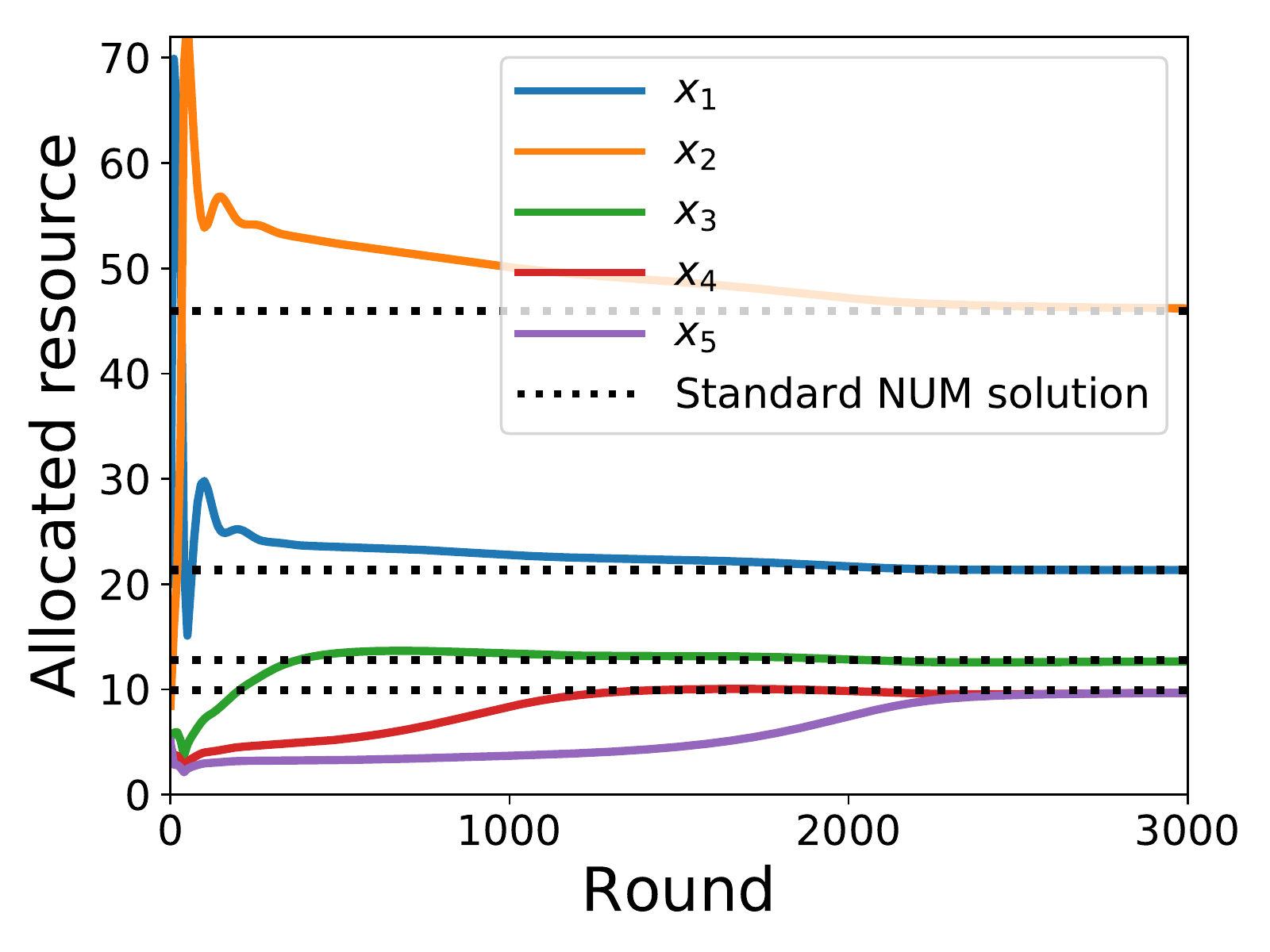}
\includegraphics[width=5cm]{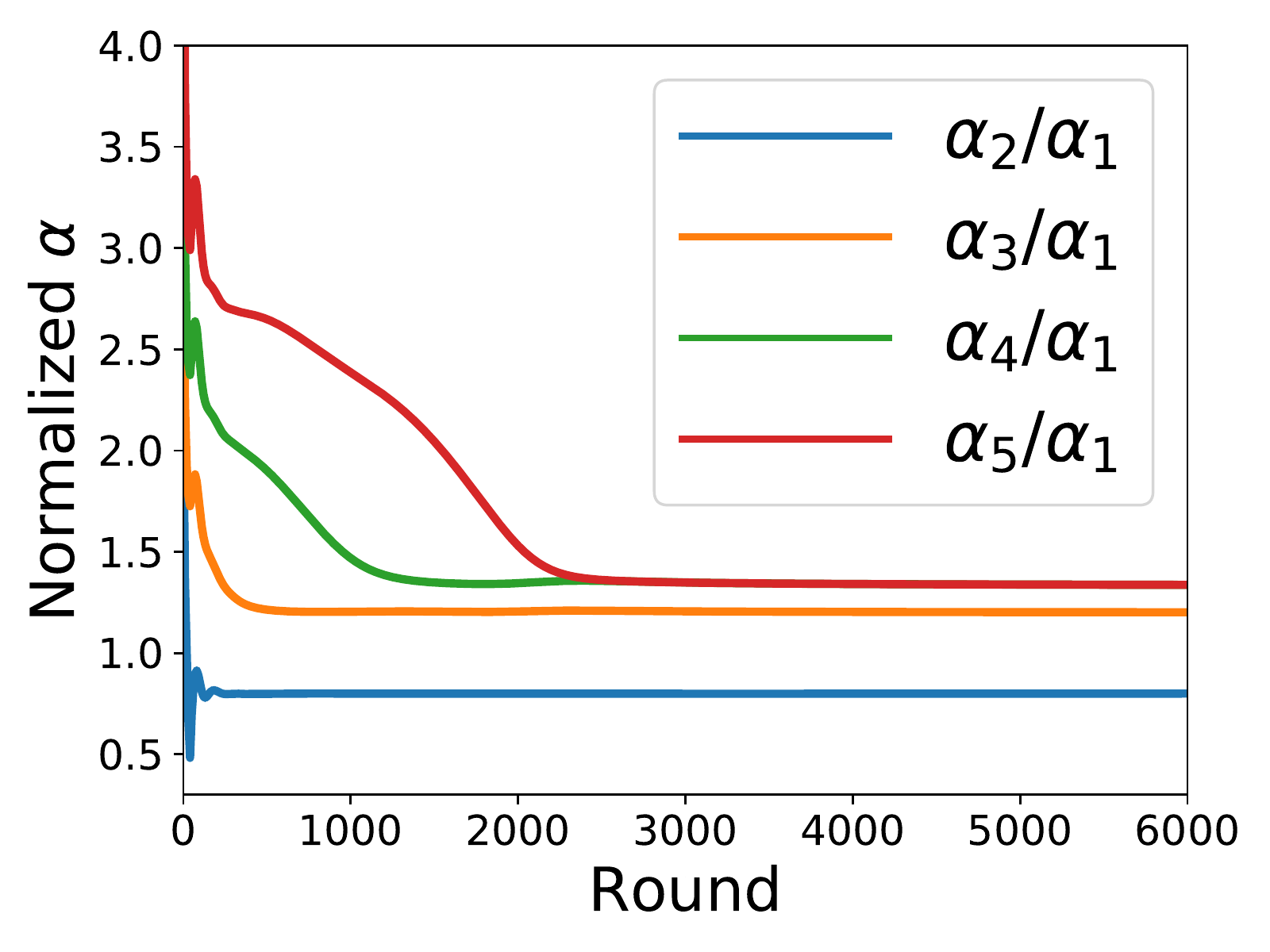}
\caption{Network utility maximization via our proposed bilevel solver DBiNUM in a $5$-user setting. Left plot: total underlying utility $\Psi$ v.s.~\# of rounds; middle plot: allocated resource v.s.~\# of rounds; right plot: normalized $\alpha$ v.s.~\# of rounds.}
\label{exp:simu_2}
  \end{figure*}
  
\section{Discussion on Lower-Level Method} 
Our method can be regarded as adding a top-level procedure over a lower-level standard network resource allocation process to improve the overall network utility. In this section, we discuss the impact of the lower-level procedure on our convergence analysis. 

As shown in \Cref{alg:bio}, the lower-level procedure adopts a distributed primal solution (see~\cite{srikant2013communication}) given by $\mathbf{x}^*(\bm{\alpha}) =\argmax_{\bm{x}}\Phi(\mathbf{x};\bm{\alpha}) = \sum_{r=1}^n \big(U_r(x_r;\alpha_r) -\frac{\epsilon x_r^2}{2}\big)- \sum_{l\in\mathcal{L}}B_l\big(\sum_{i:l\in\mathcal{L}_i}x_i\big) $, as given in \cref{eq:biobj}. To solve this objective function with given $\alpha_r$, each user first computes the gradient information $\nabla_x U_r(x_r;\alpha_r) - \epsilon x_r-\sum_{l\in \mathcal{L}_r}\nabla B_l(\sum_{i:l\in\mathcal{L}_i}x_i)$ using the information from his neighbors with shared links, and then run simple gradient-based updates. It has been shown in Propositions~\ref{prop:sc} and \ref{prop:lipschitz_Phi} that the lower-level function $\Phi(\bm{x};\bm{\alpha})$  is strongly-convex and smooth w.r.t.~$\bm{x}$, respectively. Then, based on the results for smooth convex optimization~\citep{nesterov2018smooth}, it can be shown that  a simple gradient ascent method can find the optimal maximizer with a sublinear rate. In other words, we can find a $\delta_\Phi$-accurate solution $\widehat x_{k,r}$ at  the $k^{th}$ iteration in finite steps. The accelerated gradient methods such as Nesterov acceleration can also be applied here to achieve a faster linear convergence rate. 

In reality, there exist various delays such as forward delay $T_f$ from the source to the target link and the backward delay $T_b$ for certain feedback to the source.  By choosing the stepsize inversely proportional to the maximum delay over the network, we enable to establish the asymptotic stability of the lower-level process (see Section 2.6 in  \cite{srikant2013communication})  as well as a nonasymptotic convergence guarantee (see \cite{lian2017can}). Thus, as long as we execute a sufficiently long time for this lower-level process, we can obtain a desired $\delta_\phi$-accurate solution.

\section{Simulation Studies}
\subsection{Validation of Bilevel Objective function}\label{exp_first}
In this section, we conduct experiments to underpin \Cref{th:case_study} to demonstrate that our bilevel optimization based approach in \Cref{alg:bio} recovers the standard network utility maximization solution with known utility functions. We consider a a single-link multi-user setting as in \Cref{se:bi2single}, where $n$ users transmit their package in a single communication link with capacity $P$. We consider the following problem setup. 
\begin{align*}
&\max_{\bm{\alpha}\in\mathcal{A}}\;\; \Psi(\bm{\alpha})=\sum_{r=1}^n\frac{x_r^*(\bm{\alpha})^{1-\widetilde \alpha_r}}{1-\widetilde \alpha_r}, \nonumber
\\& x_1^*(\bm{\alpha}),...,x_n^*(\bm{\alpha}) =\argmax_{x_r > 0, \sum_{r=1}^n x_r \leq P} \sum_{r=1}^n\frac{x_r^{1- \alpha_r}}{1- \alpha_r} 
-B\Big(\sum_{i=1}^nx_i\Big) 
\end{align*}
where we choose the log barrier regularization function $B(x) = -\tau \log(P-x)$ with a parameter $\tau$. For the lower-level problem, we use a simple $T$-step gradient ascent method with stepsize $\lambda$ to obtain good estimates $\widehat x_{k,r},r=1,...,n$ at each round $k$. For the constraint set, we choose $\mathcal{A}:=\{0.001<a_r\leq 100, r=1,...,n\}$ to ensure the boundedness of $\bm{\alpha}$.

\vspace{0.2cm}
\noindent {\bf Hyperparameter selection.} We choose the hyperparameters  $\lambda,\eta,\beta$ and $\tau$ from the candidate set $\{10^{-t},t=-4,-3.-2,-1,0,1,2,3,4\}$, 
 and set a large inner-loop iteration number $T$ from $\{10^{t},t=3,4,5\}$ to ensure a high-accuracy lower-level solution at each round. For all experiments, we choose the link capacity $P=100$. For the experiment in \Cref{exp:simu_1}, we consider a $3$-user setting with $n=3$, where we set $\widetilde\alpha_1=\frac{1}{2}$, $\widetilde\alpha_2=\frac{2}{3}$ and $\widetilde\alpha_3=\frac{2}{3}$. For the experiment in \Cref{exp:simu_2}, we consider a $5$-user setting with $n=3$, where we set $\widetilde\alpha_1=\frac{1}{2}$, $\widetilde\alpha_2=\frac{2}{5}$, $\widetilde\alpha_3=\frac{3}{5}$, $\widetilde\alpha_4=\frac{2}{3}$, $\widetilde\alpha_5=\frac{2}{3}$.

   \vspace{0.2cm}
\noindent {\bf Results.}  It can be seen from the left plot in \Cref{exp:simu_1} that the total underlying utility achieved by our proposed DBiNUM increases with the number of rounds, and converges to the standard NUM solution $31.77$. From the middle plot in in \Cref{exp:simu_1} , it is shown that under the choice of $\widetilde \alpha_1,\widetilde \alpha_2,\widetilde \alpha_3=\frac{1}{2},\frac{2}{3},\frac{2}{3}$ for the underlying utility functions, $x_1$ converges to the standard NUM solution $57.9$, and $x_2$ and $x_3$ converge to the same solution $20.99$ due to the identical underlying utility function with $\widetilde \alpha_2=\widetilde \alpha_3=\frac{2}{3}$. This validates our results in \Cref{th:case_study}, where we show that the bilevel solutions $x_r^*(\bm{\alpha}^*),r=1,...,n$ recover the standard NUM solution. The same observation can be made for the $5$-user case, where users $4, 5$ converges to the lowest $9.9$ due to the largest $\widetilde \alpha_4=\widetilde \alpha_5=\frac{2}{3}$, and user $2$ converges to the largest $45.9$ due to the smallest  $\widetilde \alpha_2=\frac{2}{5}$  (note that larger $\alpha$ means lower increase rate at larger $x$ and hence a smaller allocated resource). From the right plots in \Cref{exp:simu_1} and \Cref{exp:simu_2}, since the global solution $\bm{\alpha}$ is not unique, we plot the normalized solution $\alpha_i/\alpha_1,i=2,3,\cdots$. It can be clearly seen that each normalized solution converges after some rounds. 

 \vspace{-0.2cm}
 \begin{figure}[ht]
\centering
\includegraphics[width=9cm]{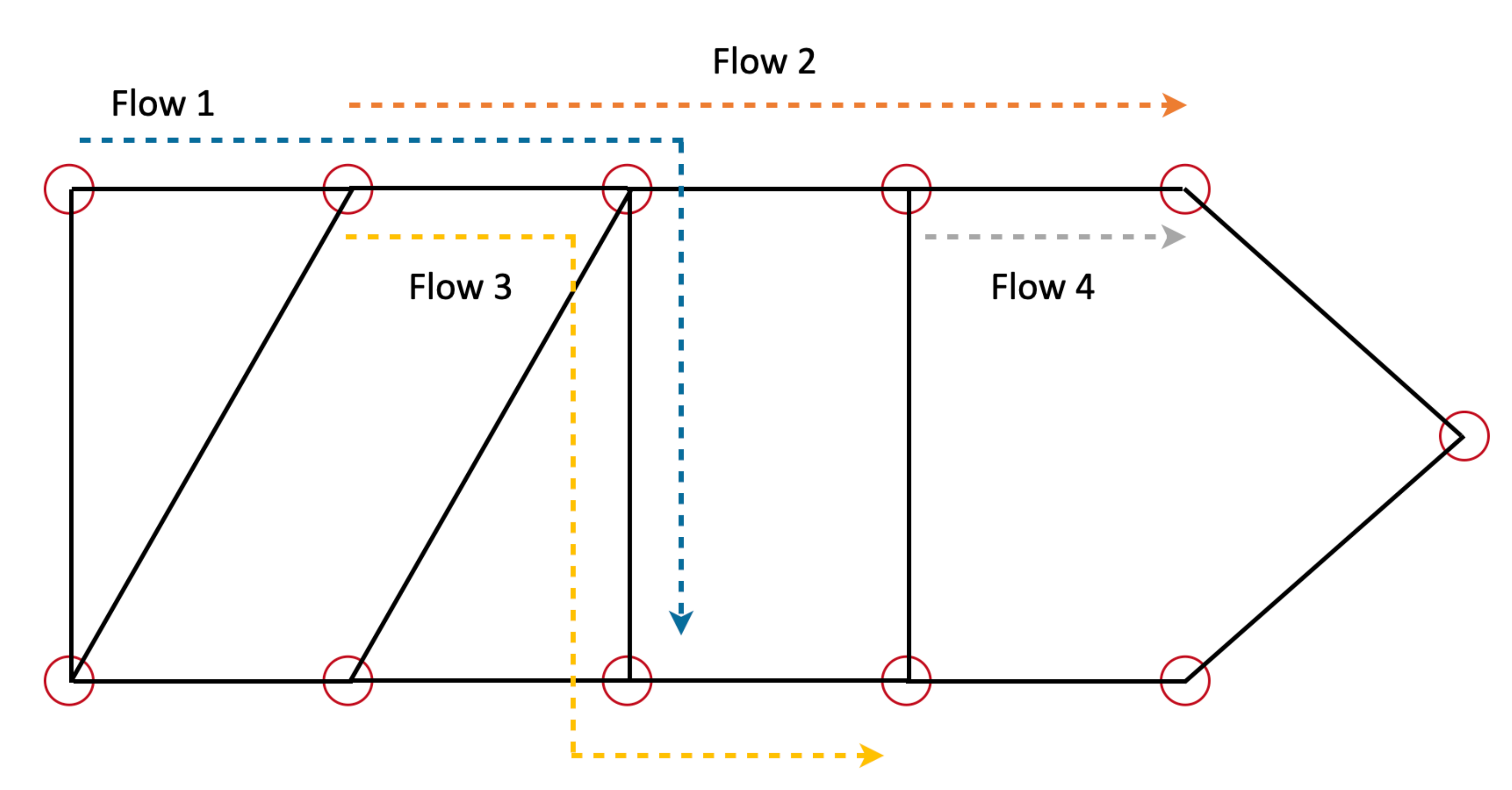}
\caption{Abilene network with four transmission flows.}
\label{fig:abilene}
\vspace{-0.2cm}
  \end{figure}
\subsection{Simulation over Real-World Networks}
In this section, we consider a real-world network, Abilene network, whose topology is shown in \Cref{fig:abilene}. 
Following the setup in \cite{fu2022learning}, 
this network contains four data transmission flows with distinct underlying utilities, where flow $1$ has a quadratic utility $a_1x^2$, flow $2$ has a square root utility $a_2\sqrt{x+b_2}-a_2\sqrt{x}$, flow $3$ has a log utility $a_3\log(b_3+1)$, and flow $4$ uses either an $\alpha$-fairness $\frac{x^{1-a_4}}{1-a_4}$ or s-shape utility~\citep{tversky1992advances} $x^{a_4}\bm{1}_{(x\geq 0)}-b_4(-x)^{a_4}\bm{1}_{(x<0)}$ ($\bm{1}_{(\cdot)}$ is the indicator function). For the bilevel objective function in \cref{eq:biobj}, we choose a log barrier regularization function $B(x)=-\tau\log(P-x)$ with a capacity $P$. 
Similarly to the setup in \Cref{exp_first}, we use a simple $T$-step gradient ascent method with stepsize $\lambda$ for the lower-level problem. For the constraint set, we choose $\mathcal{A}:=\{1.01<a_r\leq 100, r=1,...,n\}$ to ensure the boundedness of $\bm{\alpha}$.

\begin{figure}[ht]
\centering
\includegraphics[width=6cm]{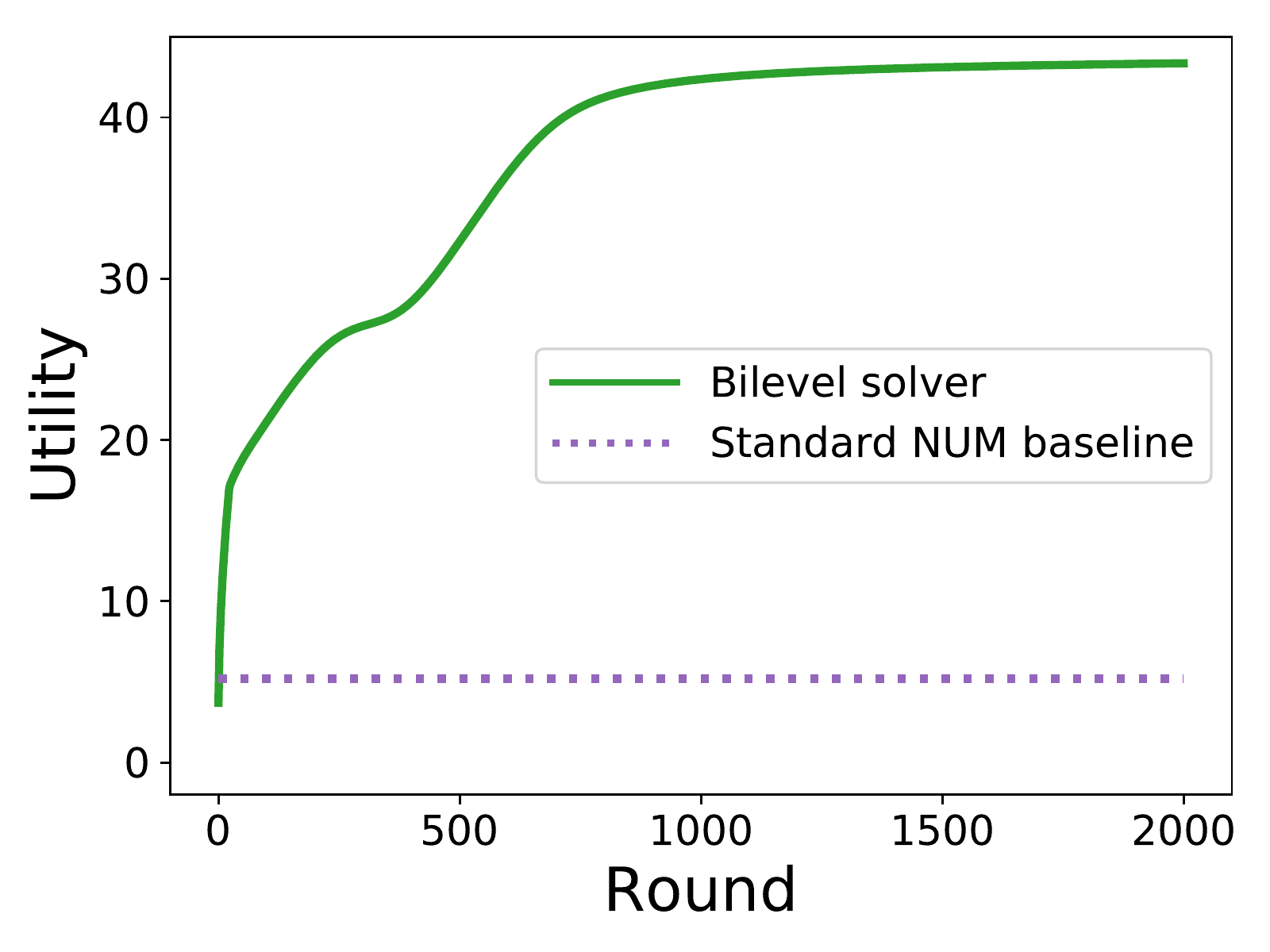}
\includegraphics[width=6cm]{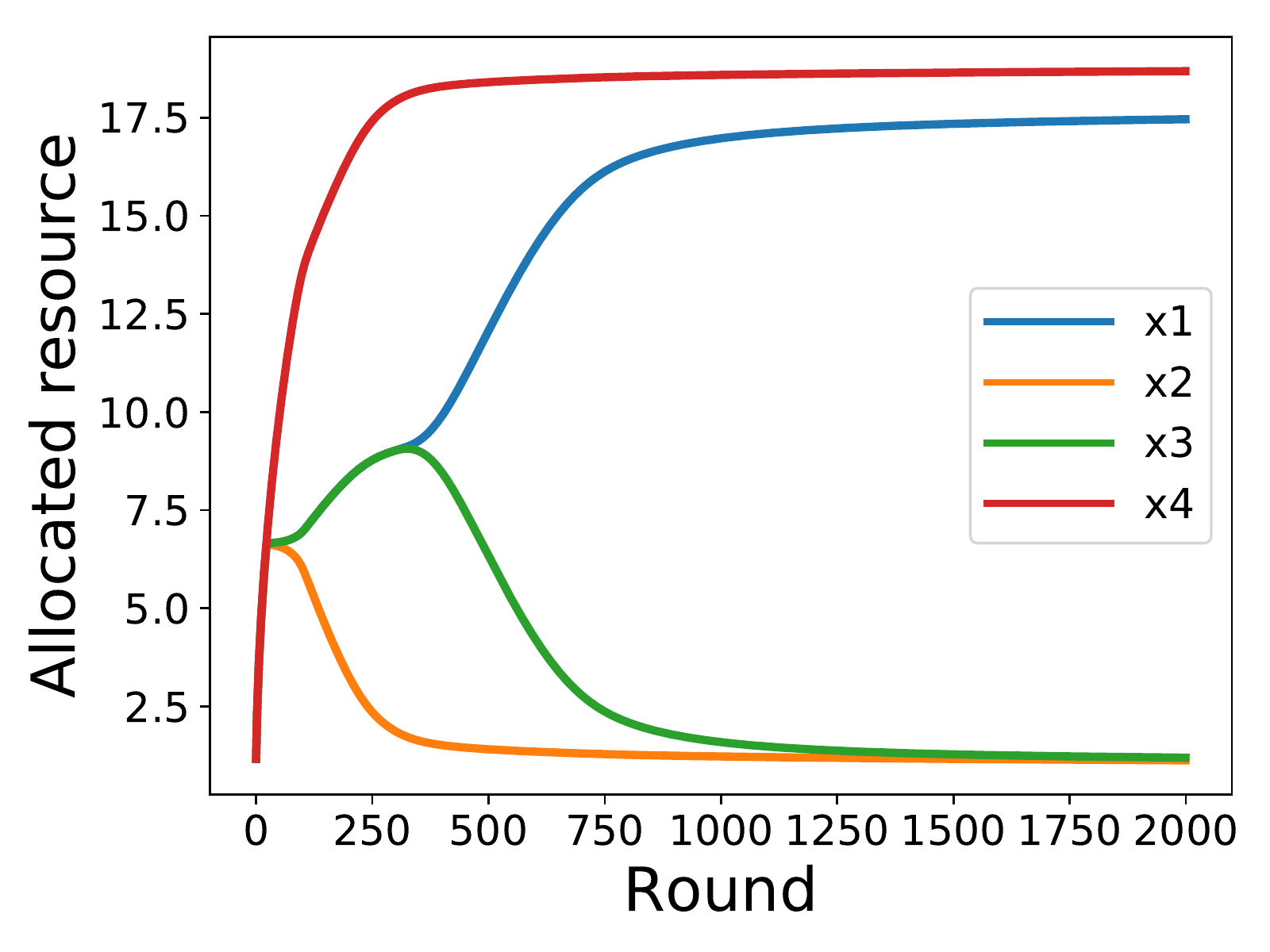}
\caption{Network utility maximization via DBiNUM in the Abilene network with $(a_1,a_2,b_2,a_3,b_3,a_4,b_4)=(0.1,5,0.4,4,1,0.8,0.2)$. Left plot: total underlying utility $\Psi$ v.s.~\# of rounds; right plot: resource v.s.~\# of rounds.}
\label{exp:simu_real}
  \end{figure}

\begin{figure}[ht]
\centering
\includegraphics[width=6cm]{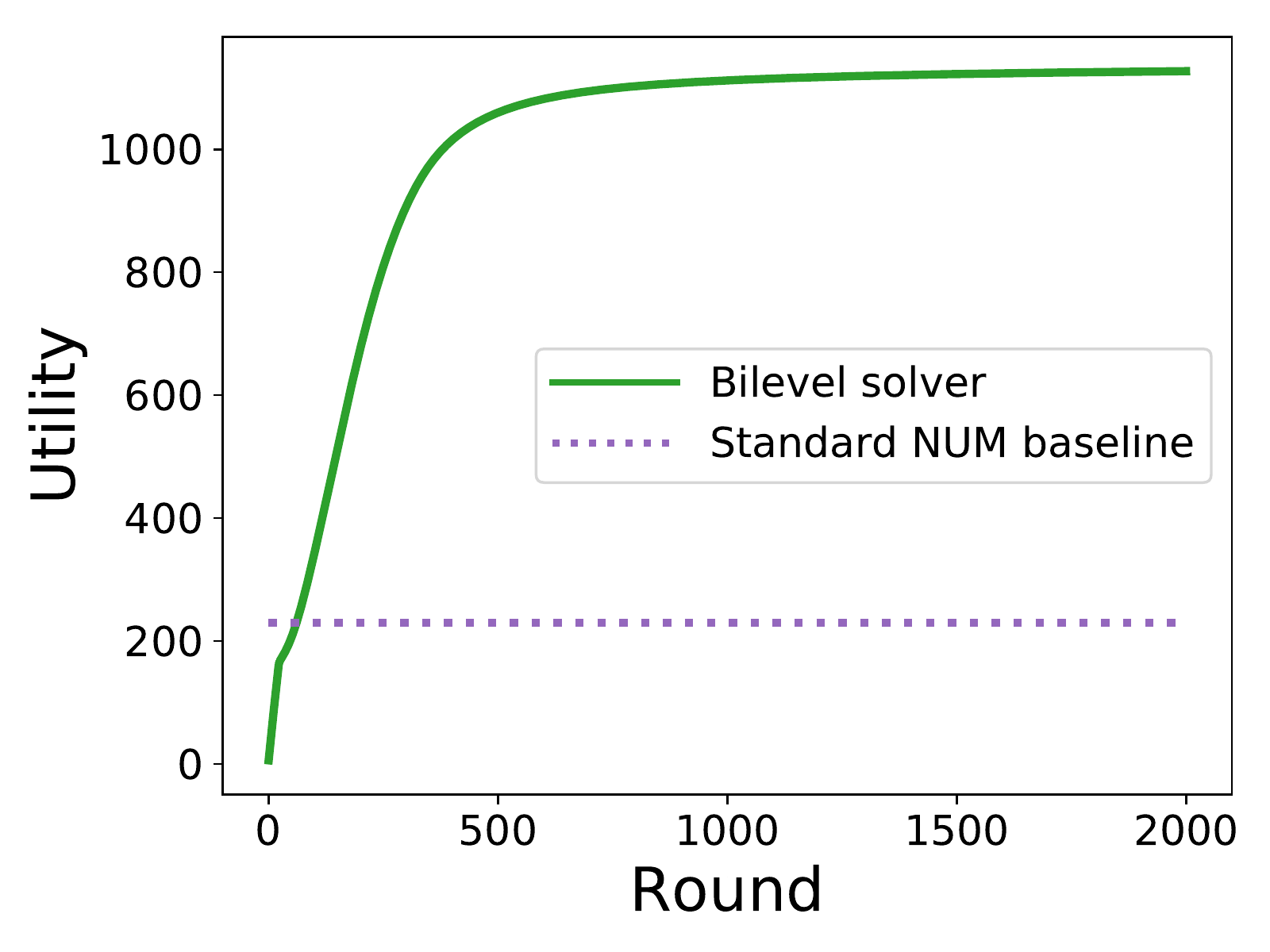}
\includegraphics[width=6cm]{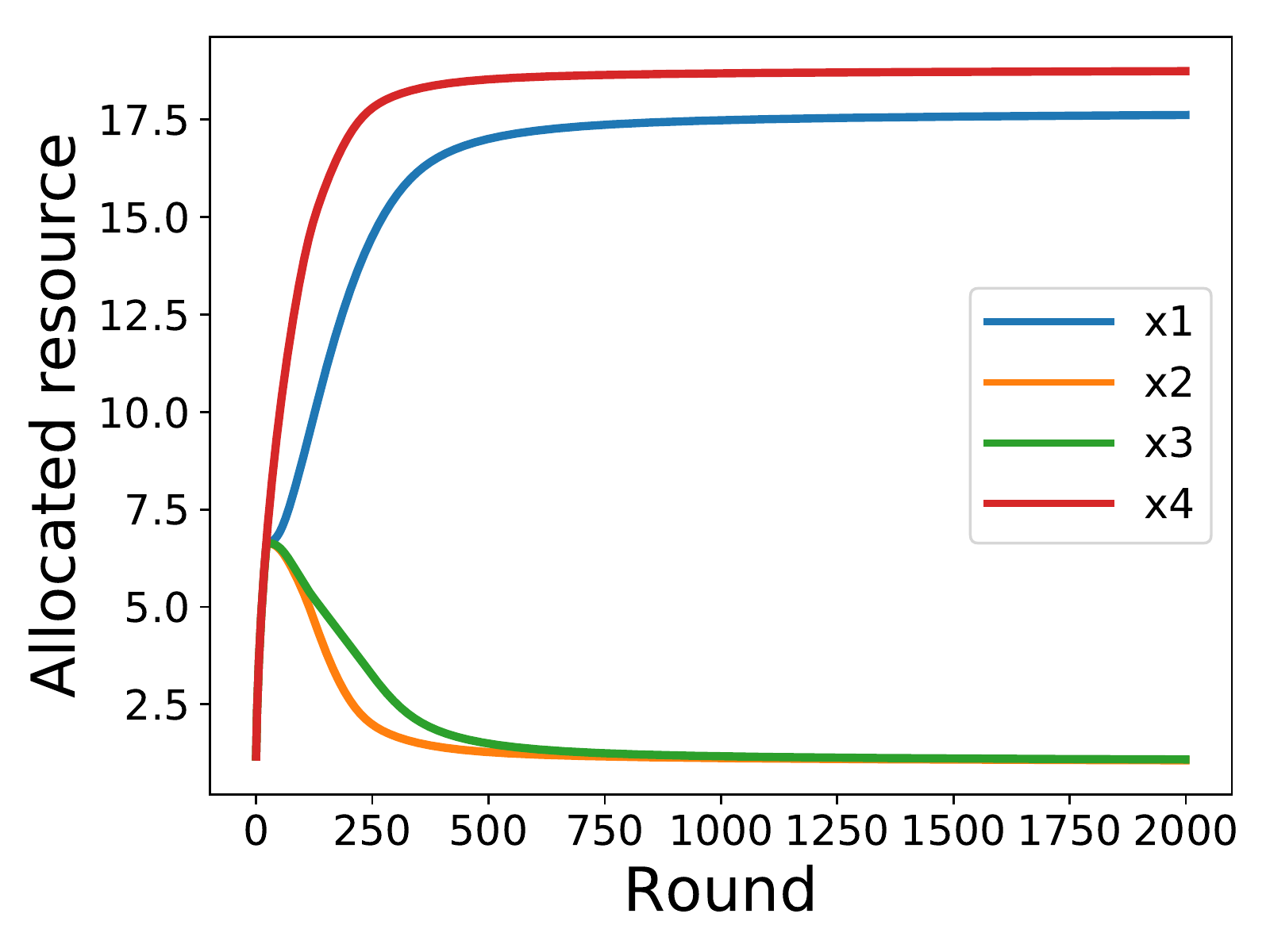}
\caption{Network utility maximization via DBiNUM in the Abilene network with $(a_1,a_2,b_2,a_3,b_3,a_4,b_4)=(3,0.5,0.2,0.5,2,1.8,2)$. Left plot: total underlying utility $\Psi$ v.s.~\# of rounds; right plot: resource v.s.~\# of rounds.}
\label{exp:simu_real2}
  \end{figure}

 \vspace{0.2cm}
\noindent {\bf Hyperparameter setting.}
For the regularization function $B(\cdot)$, we choose the constant $\tau=0.01$ and set the capacity $P=20$ for each link. The stepsizes $\lambda,\eta,\beta$ are chosen from $\{10^{t},t=-3,-2,-1,0,1,2,3\}$ to ensure the convergence. 
For the experiment in \Cref{exp:simu_real}, we set $a_1=0.1,a_2=5,b_2=0.4,a_3=4,b_3=1,a_4=0.8$ and $b_4=0.2$.
For the experiment in \Cref{exp:simu_real2}, we set $a_1=3,a_2=0.5,b_2=0.2,a_3=0.5,b_3=2,a_4=1.8$ and $b_4=2$. For both experiments, the baseline is the standard NUM solution, where each user has an $\alpha$-fairness utility function with $\alpha=2$.

  \vspace{-0.2cm}
  
  \vspace{0.2cm}
\noindent {\bf Results.} It can be seen from the left plots in \Cref{exp:simu_real} and \Cref{exp:simu_real} that our bilevel optimization method iteratively increases the underlying network utility, and greatly outperform the standard NUM baseline. For example, in \Cref{exp:simu_real}, our DBiNUM method converges to a utility of $1127.06$, which is much higher than the baseline $229.40$. The same improvement can be observed from \Cref{exp:simu_real}. This demonstrate the effectiveness of our bilevel optimization process in increasing the total underlying network utility.

\section{Conclusion and Future Work}
In this paper, we provide a novel distributed bilevel approach for network utility maximization with unknown user utility functions. Our method iteratively improves the underlying total utility based on the user feedback. Theoretically, we analyze the convergence rate of the proposed method, and show that it also recovers the standard solutions when the utility functions are known. We anticipate that our proposed theory and algorithms can motivate the design of  feasible resource allocation protocol to support distributed AI in dynamic, heterogeneous wireless networks. We also anticipate that our results will promote the development and application of  distributed bilevel optimization in the resource allocation over the communication networks.

\newpage
\appendix
\noindent{\bf \LARGE Appendix}
\section{Proof of \Cref{th:case_study}}

\allowdisplaybreaks
Let us first compute the lower-level solution $x_r^*(\bm{\alpha})$ of \cref{bilevel:case_form}. Based on the the standard analysis in \cite{srikant2013communication}, it is shown that the solutions satisfy the equality that $\sum_{r=1}^{n}x_r=P$, which implies that,  for any  index $m$,  $x_m=P-\sum_{i\neq m}x_i$. Then, 
the solutions can be obtained by setting the derive of the objective w.r.t.~$x_j$ ($j\neq m$) to be $0$, as shown below.  
\begin{align*}
\frac{\partial(\sum_{i\neq m} \frac{x_i^{1-\alpha_i}}{1-\alpha_i}+ \frac{(P-\sum_{i\neq m}x_i)^{1-\alpha_m}}{1-\alpha_m})}{\partial x_j}= x_j^{-\alpha_j} - \big(P-\sum_{i\neq m}x_i\big)^{-\alpha_m} = 0,
\end{align*}
which further yields $x_j^{-\alpha_j}=x_m^{-\alpha_m}$ for any $j\neq m$. Then, combining this relationship with the equality $\sum_{i=1}^nx_i=P$, we have $x_r^*(\bm{\alpha}),r=1,...,n$ satisfy 
\begin{align}\label{eq:lower_solutions}
\sum_{i=1}^nx_i^*(\bm{\alpha})=P,\quad x_1^*(\bm{\alpha})^{-\alpha_1}=\cdots=x_n^*(\bm{\alpha})^{-\alpha_n}.
\end{align}
Next, we derive the solutions of $\bm{\alpha}^*$ from~\cref{bilevel:case_form}. 
From \cref{eq:lower_solutions}, we have 
\begin{align}\label{eq:sum_P}
\sum_{t=1}^n (x_i^*)^{\frac{\alpha_i}{\alpha_t}} = P, 
\end{align}
where we omit $\bm{\alpha}$ for each $x_i^*$ to simplify notations. Then, for $i\neq j$, we derive the derivative $\frac{\partial x^*_i}{\partial \alpha_j}$  through 
setting the derivative of \cref{eq:sum_P} w.r.t.~$\alpha_j$ to be $0$ via implicit differentiation, as shown below.    
\begin{align*}
\sum_{t=1}^n\frac{\alpha_i}{\alpha_t} (x_i^{*})^{\frac{\alpha_i}{\alpha_t} -1}\frac{\partial x^*_i}{\partial \alpha_j} - (x_i^*)^{\frac{\alpha_i}{\alpha_j}}\frac{\alpha_i}{\alpha_j^2}\ln x_i^*  = 0,
\end{align*} 
which, by rearranging all terms, yields  
\begin{align}\label{deriva_x_i}
\frac{\partial x^*_i}{\partial \alpha_j}  = \frac{(x_i^*)^{\frac{\alpha_i}{\alpha_j}}\frac{\alpha_i}{\alpha_j^2}\ln x_i^*}{\sum_{t=1}^n\frac{\alpha_i}{\alpha_t} (x_i^{*})^{\frac{\alpha_i}{\alpha_t} -1}}.
\end{align}
For the case when $i=j$, using an approach similar to \cref{deriva_x_i}, we have 
\begin{align}\label{eq:d_xi_ai}
\frac{\partial x^*_i}{\partial \alpha_i}  = \frac{-\sum_{t\neq i}\frac{1}{\alpha_t}(x_i^*)^{\frac{\alpha_i}{\alpha_t}}\ln x_i^* }{\sum_{t=1}^n\frac{\alpha_i}{\alpha_t} (x_i^{*})^{\frac{\alpha_i}{\alpha_t} -1}}.
\end{align}
Based on the property of the derivatives we obtain in \cref{deriva_x_i} and \cref{eq:d_xi_ai}, we next derive the optimal solution $\bm{\alpha}^*$ and the resulting resource allocation $x_i^*(\bm{\alpha}^*)$. Taking the derivative of the total upper-level objective $\Psi(\bm{\alpha})=\sum_{i=1}^n\frac{x_r^*(\bm{\alpha})^{1-\widetilde \alpha_r}}{1-\widetilde \alpha_r}$ w.r.t.~$\alpha_j$ yields
\begin{align*}
\frac{\partial \Psi(\bm{\alpha})}{\partial \alpha_j} = \sum_{i\neq j} (x_i^*)^{-\widetilde\alpha_i}\frac{\partial x_i^*}{\partial \alpha_j} + (x_j^*)^{-\widetilde\alpha_j} \frac{\partial x_j^*}{\partial \alpha_j},
\end{align*}
which, combined with \cref{deriva_x_i} and \cref{eq:d_xi_ai}, yields
\begin{align}\label{eq:phi_a_grad}
\frac{\partial \Psi(\bm{\alpha})}{\partial \alpha_j} = \sum_{i\neq j}(x_i^*)^{-\widetilde\alpha_i}\frac{(x_i^*)^{\frac{\alpha_i}{\alpha_j}}\frac{\alpha_i}{\alpha_j^2}\ln x_i^*}{\sum_{t=1}^n\frac{\alpha_i}{\alpha_t} (x_i^{*})^{\frac{\alpha_i}{\alpha_t} -1}} - (x_j^*)^{-\widetilde\alpha_j}\frac{\sum_{i\neq j}\frac{1}{\alpha_i}(x_j^*)^{\frac{\alpha_j}{\alpha_i}}\ln x_j^* }{\sum_{t=1}^n\frac{\alpha_j}{\alpha_t} (x_j^{*})^{\frac{\alpha_j}{\alpha_t} -1}}.
\end{align}
For the right hand side of \cref{eq:phi_a_grad}, note that  
\begin{align}\label{eq:term_relation}
\frac{\frac{1}{\alpha_i}(x_j^*)^{\frac{\alpha_j}{\alpha_i}}\ln x_j^* }{\sum_{t=1}^n\frac{\alpha_j}{\alpha_t} (x_j^{*})^{\frac{\alpha_j}{\alpha_t} -1}} \overset{(i)}= \frac{\frac{1}{\alpha_j}x_j^*\ln x_j^*}{\sum_{t=1}^n\frac{\alpha_i}{\alpha_t}(x_j^*)^{\alpha_j(\frac{1}{\alpha_t}-\frac{1}{\alpha_i})}} \overset{(ii)}= \frac{\frac{1}{\alpha_j}x_j^*\ln x_j^*}{\sum_{t=1}^n\frac{\alpha_i}{\alpha_t}(x_i^*)^{\frac{\alpha_i}{\alpha_t}-1}} \overset{(iii)}=\frac{\frac{\alpha_i}{\alpha_j^2}(x_i^*)^{\frac{\alpha_i}{\alpha_j}}\ln x_i^*}{\sum_{t=1}^n\frac{\alpha_i}{\alpha_t}(x_i^*)^{\frac{\alpha_i}{\alpha_t}-1}},
\end{align}
where $(i)$ follows by dividing the upper and lower sides by $\frac{\alpha_j}{\alpha_i}(x_j^*)^{\frac{\alpha_j}{\alpha_i}-1}$, $(ii)$ follows because $(x_j^*)^{\alpha_j}=(x_i^*)^{\alpha_i}$ (see \cref{eq:lower_solutions}), and $(iii)$ follows because $x_j^*=(x_i^*)^{\frac{\alpha_i}{\alpha_j}}$. Then, incorporate  \cref{eq:term_relation}
into \cref{eq:phi_a_grad} yields
\begin{align}\label{eq:D_zero}
\frac{\partial \Psi(\bm{\alpha})}{\partial \alpha_j} =\frac{1}{\alpha_j}\ln x_j^*\sum_{i\neq j} ((x_i^*)^{-\widetilde\alpha_i}-(x_j^*)^{-\widetilde\alpha_j}) \frac{(x_i^*)^{\frac{\alpha_i}{\alpha_j}}}{\sum_{t=1}^n\frac{\alpha_i}{\alpha_t}(x_i^*)^{\frac{\alpha_i}{\alpha_t}-1}}.
\end{align}
We next consider two cases $P\neq n$ and $P=n$, separately. 

\vspace{0.1cm}
\noindent {\bf For $P\neq n$ case}:
\vspace{0.1cm}

Note that $x_j^*\neq 1$. Otherwise, from  \cref{eq:sum_P}, we have $n=P$, which contradicts the condition that $P\neq n$. Then,  we derive the optimal solution $\bm{\alpha}$ by setting \cref{eq:D_zero} to be $0$. This gives 
\begin{align}\label{eq:make_zero_sum}
\sum_{i\neq j} ((x_i^*)^{-\widetilde\alpha_i}-(x_j^*)^{-\widetilde\alpha_j}) \frac{(x_i^*)^{\frac{\alpha^*_i}{\alpha^*_j}}}{\sum_{t=1}^n\frac{\alpha^*_i}{\alpha^*_t}(x_i^*)^{\frac{\alpha^*_i}{\alpha^*_t}-1}} = 0
\end{align}
Let $j$ be such that $(x_j^*)^{-\widetilde \alpha_{j}}\leq(x_i^*)^{-\widetilde \alpha_{i}}$ for any $i=1,...,n$, which, combined with \cref{eq:make_zero_sum} and $x_i^*>0$, yields
\begin{align}\label{eq:n_relation}
(x_1^*)^{-\widetilde\alpha_1}=\cdots=(x_n^*)^{-\widetilde\alpha_n}.
\end{align}
Combining the above \cref{eq:n_relation} with the relationship $(x_1^*)^{-\alpha_1}=\cdots=(x_n^*)^{-\alpha_n}$ in \cref{eq:lower_solutions} and the constraint $\bm{\alpha}\in\mathcal{A}$, the solution $\bm{\alpha}^*\in\argmax_{\bm{\alpha}\in\mathcal{A}}\Psi(\bm{\alpha})$ satisfies that $\alpha_r^*=c\widetilde \alpha_r$  with $0<c\leq \frac{b}{\max_r(\widetilde \alpha_r)}$ for $r=1,...,n$. Next, we show that the resulting $\bm{x}^*(\bm{\alpha}^*)$ recovers the solution of the conventional network maximization problem in \cref{example:test_ori}. To see this, 
combining \cref{eq:n_relation} with the relationship $\sum_{i=1}^nx_i^*=P$ in \cref{eq:lower_solutions} yields that 
each $x_i^*$ satisfies $\sum_{t=1}^n (x_i^*)^{\frac{\widetilde\alpha_i}{\widetilde \alpha_t}} = P$, which can be verified to be the solution of \cref{example:test_ori}. 

\vspace{0.1cm}
\noindent {\bf For $P=n$ case}:
\vspace{0.1cm}

In this case, letting the derivative in \cref{eq:D_zero} to be $0$, it can be seen that if there exists at least one $x_j^*=1$, based on \cref{eq:lower_solutions}, all $x_1^*,...,x_n^*$ equal to $1$. Otherwise, using an approach similar to the above $P\neq n$ case, we have $\sum_{t=1}^n (x_i^*)^{\frac{\widetilde\alpha_i}{\widetilde \alpha_t}} = P$. Let $i_0$ be such that $\widetilde \alpha_{i_0}:=\max_{i}\widetilde \alpha_i$. Then, the equation $\sum_{t=1}^n (x_{i_0}^*)^{\frac{\widetilde\alpha_{i_0}}{\widetilde \alpha_t}} = P=n$ implies that $x_{i_0}^*=1$. Then, by \cref{eq:lower_solutions}, we also have the conclusion that all $x_1^*,...,x_n^*$ equal to $1$. In sum, in this $P=n$ case, we have the solution given by $x_1^*=\cdots=x_n^*=1$. Note that this also recovers the solution of \cref{example:test_ori} in the case of  $P=n$. 

\vspace{0.1cm}
Then, combining the above two cases completes the proof. 

\section{Proof of \Cref{prop:hyperG}}
First, based on the optimality of $\bm{x}^*(\bm{\alpha})$ in \cref{eq:biobj}, we have 
\begin{align}\label{eq:optimalityPhi}
\nabla_{\bm{x}}\Phi(\mathbf{x}^*(\bm{\alpha});\bm{\alpha})=0.
\end{align} 
Note that $\mathbf{x}^*(\bm{\alpha})$ is unique due to the strong-concavity of $\Phi(\cdot\,;\bm{\alpha})$ and $\Phi(\cdot\,;\cdot)$ is twice differentiable. Therefore, applying implicit differentiation to \cref{eq:optimalityPhi} yields 
\begin{align*}
\frac{\partial \mathbf{x}^*(\bm{\alpha})}{\partial \bm{\alpha}}\nabla_{\bm{x}}^2\Phi(\mathbf{x}^*(\bm{\alpha});\bm{\alpha}) + \nabla_{\bm{\alpha}}\nabla_{\bm{x}}\Phi(\mathbf{x}^*(\bm{\alpha});\bm{\alpha})  = 0,
\end{align*}
which, in conjunction with the fact that $\nabla_{\bm{x}}^2\Phi(\mathbf{x}^*(\bm{\alpha});\bm{\alpha})$ is invertible,  
 further yields 
 \begin{align}\label{eq:hinverse}
 \frac{\partial \mathbf{x}^*(\bm{\alpha})}{\partial \bm{\alpha}}=- \nabla_{\bm{\alpha}}\nabla_{\bm{x}}\Phi(\mathbf{x}^*(\bm{\alpha});\bm{\alpha}) \big( \nabla_{\bm{x}}^2\Phi(\mathbf{x}^*(\bm{\alpha});\bm{\alpha}) \big)^{-1}. 
\end{align}
The forms of $\nabla_{\bm{\alpha}}\nabla_{\bm{x}}\Phi(\mathbf{x}^*(\bm{\alpha});\bm{\alpha}) $ and $\nabla_{\bm{x}}^2\Phi(\mathbf{x}^*(\bm{\alpha});\bm{\alpha})$ in \cref{eq:hessian_form} can be proved based on the explicit forms of  of $\Phi(\cdot\,;\cdot)$ in \cref{eq:biobj}. Furthermore, taking the derivative of the upper-level function $\Psi(\bm{\alpha})=\sum_{r=1}^n \widetilde U_r(x_r^*(\alpha_r))$ w.r.t.~$\bm{\alpha}$, and using the chain rule, we have 
\begin{align*}
\frac{\Phi(\bm{\alpha})}{\partial \bm{\alpha}}=\frac{\partial \bm{x}^*(\bm{\alpha})}{\partial \bm{\alpha}}\frac{\partial \sum_{r=1}^n \widetilde U_r(x_r^*(\alpha_r))}{\partial\bm{x}} = \frac{\partial \bm{x}^*(\bm{\alpha})}{\partial \bm{\alpha}}\big[\nabla \widetilde U_1(x_1^*),...,\nabla \widetilde U_n(x_n^*)\big]^T,
\end{align*}
which, in conjunction with \cref{eq:hinverse}, finishes the proof. 

\section{Proof of \Cref{prop:sc}}

By the definition of $\Phi(\mathbf{x};\bm{\alpha}) $ in \cref{eq:biobj}, we have that the $r^{th}$ coordinate of the gradient $\nabla_\mathbf{x}\Phi(\mathbf{x};\bm{\alpha}) $ is given by 
$\nabla_{x_r}U_r(x_r;\alpha_r) -\epsilon x_r- \sum_{l\in\mathcal{L}_r}\nabla B_l \big(\sum_{i:l\in\mathcal{L}_i}x_i\big)$. Thus, for any $\bm{\alpha}\in\mathcal{A}$ and  $\mathbf{x},\mathbf{\widetilde x}\in\mathcal{X}$, we have
\begin{align*}
\big\langle \nabla_\mathbf{x}&\Phi(\mathbf{x};\bm{\alpha}), \mathbf{\widetilde x}-\mathbf{x} \big\rangle \nonumber
\\=& \sum_{r=1}^n \nabla_{x_r}U_r(x_r;\alpha_r)(\widetilde x_r-x_r)  -\sum_{r=1}^n\epsilon x_r(\widetilde x_r - x_r)
- \sum_{r=1}^n\sum_{l\in\mathcal{L}_r}\nabla B_l \Big(\sum_{i:l\in\mathcal{L}_i}x_i\Big)(\widetilde x_r-x_r) \nonumber
\\=&\sum_{r=1}^n \nabla_{x_r}U_r(x_r;\alpha_r)(\widetilde x_r-x_r) -\sum_{r=1}^n\epsilon x_r(\widetilde x_r - x_r)
- \sum_{l\in\mathcal{L}}\nabla B_l \Big(\sum_{i:l\in\mathcal{L}_i}x_i\Big)\sum_{r:l\in\mathcal{L}_r}(\widetilde x_r - x_r) \nonumber
\\=&\sum_{r=1}^n \nabla_{x_r}U_r(x_r;\alpha_r)(\widetilde x_r-x_r) -\sum_{r=1}^n\epsilon x_r(\widetilde x_r - x_r)
- \sum_{l\in\mathcal{L}}\nabla B_l \Big(\sum_{i:l\in\mathcal{L}_i}x_i\Big)\Big(\sum_{i:l\in\mathcal{L}_i}\widetilde x_i - \sum_{i:l\in\mathcal{L}_i}x_i\Big)\nonumber
\\\overset{(i)}\geq &\sum_{r=1}^n\big(U_r(\widetilde x_r;\alpha_r) -U_r(x_r;\alpha_r)\big) -\frac{\epsilon}{2}\sum_{r=1}^n (\widetilde x_r^2-x_r^2)
+\frac{\epsilon}{2}\sum_{r=1}^n(\widetilde x_r-x_r)^2  \nonumber
\\&- \sum_{l\in\mathcal{L}}B_l \Big(\sum_{i:l\in\mathcal{L}_i}\widetilde x_i\Big) + \sum_{l\in\mathcal{L}}B_l \Big(\sum_{i:l\in\mathcal{L}_i}x_i\Big)+\frac{\mu}{2}\sum_{l\in\mathcal{L}}\big(\sum_{i:l\in\mathcal{L}_i}(\widetilde x_i-x_i)\big)^2\nonumber
\\=&\Phi(\mathbf{\widetilde x}) - \Phi(\mathbf{x})  + \frac{\epsilon}{2}\|\mathbf{\widetilde x}-\mathbf{x}\|^2 + \frac{\mu M_{\min}}{2}\|\bm{\widetilde x}-\bm{x}\|^2,
\end{align*} 
where $(i)$ follows because $U_r(\cdot\;;\alpha_r)$ is $\mu$-strongly-concave and $B_l(\cdot)$ is convex, and $M_{\min} = \min_{r=1,...,n}\{M_r: \text{number of links the user $r$ exclusively occupies} \}$. Then, the proof is complete.  

\section{Proof of \Cref{prop:lipschitz_Phi}}
First, based on the form of $\Phi(\bm{x};\bm{\alpha})$ in \cref{eq:biobj}, the $i^{th}$ coordinate of the gradient $\nabla_{\bm{x}}\Phi(\bm{x};\bm{\alpha})$ is given by 
$\nabla_xU_i(x_i;\alpha_i)-\sum_{l\in\mathcal{L}_i} \nabla B_l\big(\sum_{r:l\in\mathcal{L}_r}x_r\big)$,
which, by Assumption~\ref{assum:geometry}, yields, for any $\bm{x},\bm{x}^\prime\in\mathcal{X}$ and $\bm{\alpha}$,  
\begin{align}
\|\nabla_{\bm{x}}&\Phi(\bm{x};\bm{\alpha})-\nabla_{\bm{x}}\Phi(\bm{x}^\prime;\bm{\alpha})\|^2  \nonumber
\\&\leq \sum_{i=1}^n 2(\nabla_x U_i(x_i;\alpha_i) - \nabla_x U_i(x_i^\prime;\alpha_i) )^2 + 2\sum_{i=1}^n \sum_{l\in\mathcal{L}_i} \Big(   \nabla B_l\Big(\sum_{r:l\in\mathcal{L}_r}x_r\Big)-\nabla B_l\Big(\sum_{r:l\in\mathcal{L}_r}x_r^\prime\Big)        \Big)^2 \nonumber
\\&\leq 2\sum_{i=1}^n L^2_u(x_i-x_i^\prime)^2 + 2\sum_{i=1}^n\sum_{l\in\mathcal{L}_i} L_b^2 \Big(\sum_{r:l\in\mathcal{L}_r}(x_r-x_r^\prime) \Big)^2 \nonumber
\\&\leq 2L_u^2\|\bm{x}-\bm{x}^\prime\|^2 + 2nL_b^2\sum_{i=1}^n\sum_{l\in\mathcal{L}_i}\|\bm{x}-\bm{x}^\prime\|^2 \nonumber
\\&= \Big( 2L_u^2 +2n\sum_{i=1}^n\sum_{l\in\mathcal{L}_i}L_b^2 \Big) \|\bm{x}-\bm{x}^\prime\|^2,
\end{align}
which, by taking the square root at the both sides, yields the proof for the Lipschitz continuity  of $\nabla_{\bm{x}}\Phi(\cdot\,;\bm{\alpha})$. 
Based on the form of Hessian $\nabla^2_{\bm{x}} \Phi(\bm{x};\bm{\alpha})$ in \cref{eq:hessian_form}, we have, for any given vector $\bm{u}=[u_1,...,u_n]^T$,
\begin{align}
\|\nabla^2_{\bm{x}} \Phi(\bm{x};\bm{\alpha})&\bm{u} - \nabla^2_{\bm{x}} \Phi(\bm{x}^{\prime};\bm{\alpha})\bm{u}\|^2\nonumber
\\=&\sum_{i=1}^n \Big(\nabla^2_{x} U_i(x_i;\alpha_i)u_i -\nabla^2_{x} U_i(x_i^\prime;\alpha_i) u_i \nonumber
\\& -\sum_{j:\mathcal{L}_i\cap\mathcal{L}_j\neq\O}\sum_{l\in\mathcal{L}_i\cap\mathcal{L}_j} \Big(\nabla^2B_l\big(\sum_{r:l\in\mathcal{L}_r}x_r\big)-\nabla^2B_l\big(\sum_{r:l\in\mathcal{L}_r}x_r^\prime\big)\Big) u_i \Big)^2\nonumber
\\\overset{(i)}\leq&\sum_{i=1}^n 2\Big(\nabla^2_{x} U_i(x_i;\alpha_i)-\nabla^2_{x} U_i(x_i^\prime;\alpha_i)\Big)^2u_i^2 \nonumber
\\&+ \sum_{i=1}^n 2 \Big(\sum_{j:\mathcal{L}_i\cap\mathcal{L}_j\neq\O}\sum_{l\in\mathcal{L}_i\cap\mathcal{L}_j} L_b\Big | \sum_{r:l\in\mathcal{L}_r}\big(x_r-x_r^\prime\big)\Big | |u_i|  \Big)^2 \nonumber
\\\leq & \sum_{i=1}^n 2L_u^2(x_i-x
_i^\prime)^2u_i^2 + \sum_{i=1}^n 2 \Big(\sum_{r=1}^n\big |x_r-x_r^\prime\big|\sum_{j:\mathcal{L}_i\cap\mathcal{L}_j\neq\O}\sum_{l\in\mathcal{L}_i\cap\mathcal{L}_j} L_b |u_i|  \Big)^2 \nonumber
\\\leq& \Big(\sum_{i=1}^n 2L_u^2u_i^2\Big)\|\bm{x}-\bm{x}^\prime\|^2 + \sum_{i=1}^n 2 \Big(\sum_{j:\mathcal{L}_i\cap\mathcal{L}_j\neq\O}\sum_{l\in\mathcal{L}_i\cap\mathcal{L}_j} L_b |u_i|  \Big)^2 (\sum_{r=1}^n\big |x_r-x_r^\prime\big|)^2 \nonumber
\\\leq& \Big( 2L_u^2 + 2n L_b^2\max_i\Big(\sum_{j:\mathcal{L}_i\cap\mathcal{L}_j\neq\O}\sum_{l\in\mathcal{L}_i\cap\mathcal{L}_j} 1  \Big)^2 \Big)\max_{i}u_i^2\|\bm{x}-\bm{x}^\prime\|^2,
\end{align}
which proves the Lipschitz continuity of $\nabla^2_{\bm{x}} \Phi(\cdot\,;\bm{\alpha})\bm{u}$. We next show the Lipschitz continuity  of $\nabla^2_{\bm{x}} \Phi(\bm{x};\cdot)\bm{u}$. Similarly, based on the form of $\nabla^2_{\bm{x}} \Phi(\bm{x};\bm{\alpha})$ in \cref{eq:hessian_form}, we have, for any $\bm{\alpha}$ and $\bm{\alpha}^\prime$ 
\begin{align}
\|\nabla^2_{\bm{x}} \Phi(\bm{x};\bm{\alpha})\bm{u} - \nabla^2_{\bm{x}} \Phi(\bm{x};\bm{\alpha}^\prime)\bm{u}\|^2 =  &\sum_{i=1}^n (\nabla^2_{x} U_i(x_i;\alpha_i)u_i-\nabla^2_{x} U_i(x_i;\alpha_i^\prime)u_i)^2 \nonumber
\\\leq & \sum_{i=1}^n L_u^2(\alpha_i-\alpha_i^\prime)^2u_i^2\leq L_u^2\max_{i}u_i^2\|\bm{\alpha}-\bm{\alpha}^\prime\|^2.
\end{align}
Finally, we prove the Lipschitz continuity of the mixed derivative $\nabla_{\bm{\alpha}} \nabla_{\bm{x}} \Phi(\bm{x};\bm{\alpha}) \bm{u}$. Noting that $\nabla_{\bm{\alpha}} \nabla_{\bm{x}} \Phi(\bm{x};\bm{\alpha})$ is a diagonal matrix whose $i^{th}$ diagonal element is $\nabla_{\alpha}\nabla_{x}U_i(x_i;\alpha_i)$. Then, we have, for any $\bm{x},\bm{x}^\prime$ 
\begin{align}
\|\nabla_{\bm{\alpha}} \nabla_{\bm{x}} \Phi(\bm{x};\bm{\alpha}) \bm{u}-\nabla_{\bm{\alpha}} \nabla_{\bm{x}} \Phi(\bm{x}^\prime;\bm{\alpha})\bm{u} \|^2\leq \sum_{i=1}^n L_u^2(x_i-x_i^\prime)^2u_i^2\leq L_u^2\max_{i}u_i^2\|\bm{x}-\bm{x}^\prime\|^2.
\end{align}
Similarly, it can be shown that $\|\nabla_{\bm{\alpha}} \nabla_{\bm{x}} \Phi(\bm{x};\bm{\alpha}) \bm{u}-\nabla_{\bm{\alpha}} \nabla_{\bm{x}} \Phi(\bm{x};\bm{\alpha}^{\prime})\bm{u} \|^2\leq L_u^2\max_{i}u_i^2\|\bm{\alpha}-\bm{\alpha}^\prime\|^2.
$ Then, the proof is complete. 
\section{Proof of \Cref{prop:v_updates}}
  Based on the update in \cref{v:updates}, the auxiliary vector $\bm{v}_{k+1}$ can be written as   
  \begin{align}\label{eq:one_step_estimate}
\bm{v}_{k+1} =\big (I+\eta \nabla^2_{\bm{x}} \Phi(\bm{\widehat x}_k;\bm{\alpha}_k) \big)\bm{v}_{k}  - \eta \nabla\widetilde U(\bm{\widehat x}_k),
\end{align}
which can be regarded as an one-step estimate of the Hessian-inverse-vector $
\big( \nabla^2_{\bm{x}} \Phi(\bm{x}^*_k;\bm{\alpha}_k) \big)^{-1}\nabla\widetilde U(\bm{x}_k^*)$. 
Note that based on \Cref{alg:bio}, we have $x^*_{k,r}- \delta_\Phi<\widehat x_{k,r}< x^*_{k,r} + \delta_\Phi$, which, combined with $\delta<x_{k,r}^*<b$ in Assumption~\ref{assum:boundX} and $\delta_\Phi < \frac{\delta}{2}$, yields $\widehat x_{k,r}<b+\delta_\Phi<2b$ and $\widehat x_{k,r}>\delta-\delta_\Phi>\frac{\delta}{2}$ and hence $\bm{\widehat x}_{k}\in\mathcal{X}$. Also note that $\bm{x}_k^*\in\mathcal{X}$. Then, based on the update of \cref{eq:one_step_estimate}, we have  
 \begin{align*}
 \bm{v}_{k+1} - \big(\nabla^2_{\bm{x}} \Phi(\bm{\widehat x}_k;\bm{\alpha}_k)\big)^{-1}\nabla \widetilde U(\bm{\widehat x}_k) = (1+\eta\nabla^2_{\bm{x}} \Phi(\bm{\widehat x}_k;\bm{\alpha}_k))\big(  \bm{v}_{k} - \big(\nabla^2_{\bm{x}} \Phi(\bm{\widehat x}_k;\bm{\alpha}_k)\big)^{-1}\nabla \widetilde U(\bm{\widehat x}_k) \big),
 \end{align*}
 which, using the strong concavity of $\Phi(\cdot;\bm{\alpha}_k)$ established in \Cref{prop:sc}, yields 
 \begin{align}\label{eq:iterative_relation}
 \| \bm{v}_{k+1} - (\nabla^2_{\bm{x}} \Phi(\bm{\widehat x}_k;\bm{\alpha}_k))^{-1}\nabla \widetilde U(\bm{\widehat x}_k) \| \leq \big(1-\eta\mu_\Phi\big)\big\|  \bm{v}_{k} - (\nabla^2_{\bm{x}} \Phi(\bm{\widehat x}_k;\bm{\alpha}_k))^{-1}\nabla \widetilde U(\bm{\widehat x}_k) \big\|.
 \end{align}
 Note that the difference between $(\nabla^2_{\bm{x}} \Phi(\bm{\widehat x}_k;\bm{\alpha}_k))^{-1}\nabla \widetilde U(\bm{\widehat x}_k)$ and  $
( \nabla^2_{\bm{x}} \Phi(\bm{x}^*_k;\bm{\alpha}_k) )^{-1}\nabla\widetilde U(\bm{x}_k^*)$ is given by 
\begin{align}\label{eq:hessian_inv_vec_b}
\|(\nabla^2_{\bm{x}} \Phi(&\bm{\widehat x}_k;\bm{\alpha}_k))^{-1}\nabla \widetilde U(\bm{\widehat x}_k)-( \nabla^2_{\bm{x}} \Phi(\bm{x}^*_k;\bm{\alpha}_k) )^{-1}\nabla\widetilde U(\bm{x}_k^*)\| \nonumber
\\\leq&\| (\nabla^2_{\bm{x}} \Phi(\bm{\widehat x}_k;\bm{\alpha}_k) ^{-1}- \nabla^2_{\bm{x}} \Phi(\bm{x}^*_k;\bm{\alpha}_k) ^{-1})\nabla\widetilde U(\bm{x}_k^*)\| + \|\nabla^2_{\bm{x}} \Phi(\bm{\widehat x}_k;\bm{\alpha}_k)^{-1}\|\|\nabla \widetilde U(\bm{\widehat x}_k)-\nabla\widetilde U(\bm{x}_k^*)\| \nonumber
\\\overset{(i)}\leq &\frac{\sqrt{n}L_{\text{Hess}}L_u}{\mu_\Phi^2}\|\bm{\widehat x}_k- \bm{x}_k^*\|+ \frac{L_u}{\mu_\Phi}\|\bm{\widehat x}_k- \bm{x}_k^*\| \nonumber
\\\overset{(ii)}\leq & \Big(\frac{L_{\text{Hess}}L_u}{\mu_\Phi^2} +\frac{L_u}{\sqrt{n}\mu_\Phi} \Big)n\delta_\Phi,
\end{align}
 where $(i)$ follows from Assumption~\ref{assum:geometry}, \Cref{prop:sc} and \Cref{prop:lipschitz_Phi}, and $(ii)$ follows because $|\widehat x_{k,r} - x_{k,r}^*|<\delta_{\Phi}$. 
  Substituting \cref{eq:hessian_inv_vec_b} into \cref{eq:iterative_relation}, we have 
  \begin{align*}
   \| \bm{v}_{k+1} - ( \nabla^2_{\bm{x}} &\Phi(\bm{x}^*_k;\bm{\alpha}_k) )^{-1}\nabla\widetilde U(\bm{x}_k^*) \| \nonumber
   \\ \leq &\big(1-\eta\mu_\Phi\big)\big\|  \bm{v}_{k} - ( \nabla^2_{\bm{x}} \Phi(\bm{x}^*_k;\bm{\alpha}_k) )^{-1}\nabla\widetilde U(\bm{x}_k^*)  \big\|+\big(2-\eta\mu_\Phi \big)\Big(\frac{L_{\text{Hess}}L_u}{\mu_\Phi^2} +\frac{L_u}{\sqrt{n}\mu_\Phi} \Big)n\delta_\Phi, 
  \end{align*}
  which, using the Young's inequality that $(a+b)^2\leq(1+\lambda)a^2+(1+\frac{1}{\lambda})b^2$, yields
  \begin{align}\label{eq:v_iter_rela}
   \| \bm{v}_{k+1} - ( \nabla^2_{\bm{x}} &\Phi(\bm{x}^*_k;\bm{\alpha}_k) )^{-1}\nabla\widetilde U(\bm{x}_k^*) \|^2 \nonumber
   \\ \leq &\big(1+\eta\mu_\Phi \big)\big(1-\eta\mu_\Phi\big)^2\big\|  \bm{v}_{k} - ( \nabla^2_{\bm{x}} \Phi(\bm{x}^*_k;\bm{\alpha}_k) )^{-1}\nabla\widetilde U(\bm{x}_k^*)  \big\|^2 \nonumber
   \\ &+\Big(1+\frac{1}{\eta\mu_\Phi} \Big)\big(2-\eta\mu_\Phi\big)^2\Big(\frac{L_{\text{Hess}}L_u}{\mu_\Phi^2} +\frac{L_u}{\sqrt{n}\mu_\Phi} \Big)^2 n^2\delta_\Phi^2  \nonumber
   \\\leq &\big(1-\eta\mu_\Phi\big)\big\|  \bm{v}_{k} - ( \nabla^2_{\bm{x}} \Phi(\bm{x}^*_k;\bm{\alpha}_k) )^{-1}\nabla\widetilde U(\bm{x}_k^*)  \big\|^2 \nonumber
    \\ &+4\Big(1+\frac{1}{\eta\mu_\Phi} \Big)\Big(\frac{L_{\text{Hess}}L_u}{\mu_\Phi^2} +\frac{L_u}{\sqrt{n}\mu_\Phi} \Big)^2 n^2\delta_\Phi^2.
  \end{align} 
 We next bound the difference between $( \nabla^2_{\bm{x}} \Phi(\bm{x}^*_k;\bm{\alpha}_k) )^{-1}\nabla\widetilde U(\bm{x}_k^*) $ and $( \nabla^2_{\bm{x}} \Phi(\bm{x}^*_{k-1};\bm{\alpha}_{k-1}) )^{-1}\nabla\widetilde U(\bm{x}_{k-1}^*) $ in two adjacent iterations. Similarly to \cref{eq:hessian_inv_vec_b}, we have 
 \begin{align*}
\big \|( \nabla^2_{\bm{x}} \Phi(\bm{x}^*_k;&\bm{\alpha}_k) )^{-1}\nabla\widetilde U(\bm{x}_k^*) - ( \nabla^2_{\bm{x}} \Phi(\bm{x}^*_{k-1};\bm{\alpha}_{k-1}) )^{-1}\nabla\widetilde U(\bm{x}_{k-1}^*) \big\| \nonumber
\\\leq &\| (\nabla^2_{\bm{x}} \Phi(\bm{x}^*_k;\bm{\alpha}_k) ^{-1}- \nabla^2_{\bm{x}} \Phi(\bm{x}^*_{k-1};\bm{\alpha}_{k-1}) ^{-1})\nabla\widetilde U(\bm{x}_{k-1}^*)\| \nonumber
\\&+ \|\nabla^2_{\bm{x}} \Phi(\bm{x}^*_k;\bm{\alpha}_k)^{-1}\|\|\nabla \widetilde U(\bm{x}^*_k)-\nabla\widetilde U(\bm{x}_{k-1}^*)\| \nonumber
\\\leq & \frac{\sqrt{n}L_u}{\mu_\Phi^2} (L_{\text{Hess}} \|\bm{x}_k^*-\bm{x}_{k-1}\|^*+L_u\|\bm{\alpha}_k-\bm{\alpha}_{k-1}\|) + \frac{L_u}{\mu_\Phi}\|\bm{x}^*_k- \bm{x}_{k-1}^*\|, 
  \end{align*}
 which, using Lemma 2.2 in \cite{ghadimi2018approximation} that $\|\bm{x}^*_k-\bm{x}^*_{k-1}\|\leq \frac{L_{\text{grad}}}{\mu_\Phi}\|\bm{\alpha}_k-\bm{\alpha}_{k-1}\|$, yields 
 \begin{align}\label{eq:h_v_gap}
 \big \|( \nabla^2_{\bm{x}} \Phi(\bm{x}^*_k;&\bm{\alpha}_k) )^{-1}\nabla\widetilde U(\bm{x}_k^*) - ( \nabla^2_{\bm{x}} \Phi(\bm{x}^*_{k-1};\bm{\alpha}_{k-1}) )^{-1}\nabla\widetilde U(\bm{x}_{k-1}^*) \big\|  \nonumber
 \\&\leq \Big( \frac{L_{\text{grad}}}{\mu_\Phi}\Big(\frac{ L_uL_{\text{Hess}}}{\mu_\Phi^2} +\frac{L_u}{\sqrt{n}\mu_\Phi} \Big) + \frac{L_u^2}{\mu_\Phi^2} \Big) \sqrt{n}\|\bm{\alpha}_k-\bm{\alpha}_{k-1}\|.
 \end{align}
 Then, substituting \cref{eq:h_v_gap} into \cref{eq:v_iter_rela}, and using the Young's inequality, we have 
  \begin{align*}
 \| \bm{v}_{k+1} - ( \nabla^2_{\bm{x}} &\Phi(\bm{x}^*_k;\bm{\alpha}_k) )^{-1}\nabla\widetilde U(\bm{x}_k^*) \|^2  \nonumber
 \\\leq& \big(1-\eta\mu_\Phi\big)(1+\tau)\big\|  \bm{v}_{k} - ( \nabla^2_{\bm{x}} \Phi(\bm{x}^*_{k-1};\bm{\alpha}_{k-1}) )^{-1}\nabla\widetilde U(\bm{x}_{k-1}^*)  \big\|^2\nonumber
 \\ &+  \big(1-\eta\mu_\Phi\big)\Big(1+\frac{1}{\tau}\Big) \Big( \frac{L_{\text{grad}}}{\mu_\Phi}\Big(\frac{ L_uL_{\text{Hess}}}{\mu_\Phi^2} +\frac{L_u}{\sqrt{n}\mu_\Phi} \Big) + \frac{L_u^2}{\mu_\Phi^2} \Big)^2n\|\bm{\alpha}_k-\bm{\alpha}_{k-1}\|^2 \nonumber
  \\ &+4\Big(1+\frac{1}{\eta\mu_\Phi} \Big)\Big(\frac{L_{\text{Hess}}L_u}{\mu_\Phi^2} +\frac{L_u}{\sqrt{n}\mu_\Phi} \Big)^2 n^2\delta_\Phi^2,
 \end{align*}
which, by choosing $\tau=\frac{\eta\mu_\Phi}{2}$ and based on the definitions of $C_v$ and $\Delta_\Phi$, which finishes the proof. 
\section{Proof of \Cref{prop:hyperGapproximate}}
Using the form of $\nabla \Psi(\bm{\alpha}) $ in \Cref{prop:hyperG}, we have 
\begin{align}\label{eq:hyperG_err}
\big\|-\nabla_{\bm{\alpha}} \nabla_{\bm{x}} &\Phi(\bm{\widehat x}_k;\bm{\alpha}_k) \bm{v}_{k+1} -\nabla \Psi(\bm{\alpha}_k)\big\|\nonumber
\\\leq & \big\|  \nabla_{\bm{\alpha}} \nabla_{\bm{x}} \Phi(\bm{\widehat x}_k;\bm{\alpha}_k) \big(\bm{v}_{k+1} - ( \nabla^2_{\bm{x}} \Phi(\bm{x}^*_k;\bm{\alpha}_k) )^{-1}\nabla\widetilde U(\bm{x}_k^*)\big)\big\|  \nonumber
\\&+ \|(\nabla_{\bm{\alpha}} \nabla_{\bm{x}} \Phi(\bm{\widehat x}_k;\bm{\alpha}_k)- \nabla_{\bm{\alpha}} \nabla_{\bm{x}} \Phi(\bm{x}^*_k;\bm{\alpha}_k))( \nabla^2_{\bm{x}} \Phi(\bm{x}^*_k;\bm{\alpha}_k) )^{-1}\nabla\widetilde U(\bm{x}_k^*)\| \nonumber
\\\overset{(i)}\leq & \big\|  \nabla_{\bm{\alpha}} \nabla_{\bm{x}} \Phi(\bm{\widehat x}_k;\bm{\alpha}_k) \big(\bm{v}_{k+1} - ( \nabla^2_{\bm{x}} \Phi(\bm{x}^*_k;\bm{\alpha}_k) )^{-1}\nabla\widetilde U(\bm{x}_k^*)\big)\big\| + \frac{\sqrt{n}L^2_u}{\mu_\Phi}\|\bm{\widehat x}_k-\bm{x}^*_k\|
\end{align}
where $(i)$ follows from \Cref{prop:lipschitz_Phi} with $\max_i|u_i|\leq\|\bm{u}\|$. We next upper bound the first term at the right hand side of \cref{eq:hyperG_err}. Similarly to the proof of \Cref{prop:v_updates}, we have $\bm{\widehat x}_k\in\mathcal{X}$. Then, we have, for any $\bm{u}$, 
\begin{align}\label{eq:mixHv}
\|\nabla_{\bm{\alpha}} \nabla_{\bm{x}} \Phi(\bm{\widehat x}_k;\bm{\alpha}_k)\bm{u}\|^2 = 
\sum_{i=1}^n (\nabla_{\alpha}\nabla_{x}U_i(\widehat x_{k,i};\alpha_{k,i})u_i)^2\leq L_u^2\|\bm{u}\|^2.
\end{align}
Applying \cref{eq:mixHv} to \cref{eq:hyperG_err} yields 
\begin{align}\label{eq:mixHest}
\big\|-\nabla_{\bm{\alpha}} \nabla_{\bm{x}} \Phi(\bm{\widehat x}_k;\bm{\alpha}_k) \bm{v}_{k+1} - \nabla \Psi(\bm{\alpha}_k) \big\| \leq &L_u\big\|\bm{v}_{k+1} - ( \nabla^2_{\bm{x}} \Phi(\bm{x}^*_k;\bm{\alpha}_k) )^{-1}\nabla\widetilde U(\bm{x}_k^*)\big\| + \frac{nL^2_u}{\mu_\Phi}\delta_\Phi.
\end{align}
Substituting the update in~\cref{eq:alpha_updates}  into \Cref{prop:v_updates}, we have 
\begin{align*}
 \| \bm{v}_{k+1} - ( \nabla^2_{\bm{x}}& \Phi(\bm{x}^*_k;\bm{\alpha}_k) )^{-1}\nabla\widetilde U(\bm{x}_k^*) \|^2  \leq \big(1-\frac{\eta\mu_\Phi}{2}\big)\big\|  \bm{v}_{k} - ( \nabla^2_{\bm{x}} \Phi(\bm{x}^*_{k-1};\bm{\alpha}_{k-1}) )^{-1}\nabla\widetilde U(\bm{x}_{k-1}^*)  \big\|^2 \nonumber
 \\&+ C_v\beta^2 \|\beta^{-1}\big(\bm{\alpha}_{k-1} -  \mathcal{P}_{\mathcal{A}}\big\{ \bm{\alpha}_{k-1} -\beta \nabla_{\bm{\alpha}} \nabla_{\bm{x}} \Phi(\bm{\widehat x}_{k-1};\bm{\alpha}_{k-1}) \bm{v}_{k}   \big\}\big)\|^2 + \Delta_\Phi,
\end{align*}
which, in conjunction with the non-expansion of the projection and \cref{eq:mixHest}, yields 
\begin{align}\label{eq:v_error_final}
 \| \bm{v}_{k+1} - ( \nabla^2_{\bm{x}}& \Phi(\bm{x}^*_k;\bm{\alpha}_k) )^{-1}\nabla\widetilde U(\bm{x}_k^*) \|^2  \nonumber
 \\\leq & \big(1-\frac{\eta\mu_\Phi}{2}\big)\big\|  \bm{v}_{k} - ( \nabla^2_{\bm{x}} \Phi(\bm{x}^*_{k-1};\bm{\alpha}_{k-1}) )^{-1}\nabla\widetilde U(\bm{x}_{k-1}^*)  \big\|^2 \nonumber
 \\&+ 2C_v\beta^2 \|\beta^{-1}\big(\bm{\alpha}_{k-1} -  \mathcal{P}_{\mathcal{A}}\big\{ \bm{\alpha}_{k-1} +\beta \nabla \Psi(\bm{\alpha}_{k-1})\big\}\|^2 + \Delta_\Phi \nonumber
 \\&+ 2C_v\beta^2 \|-\nabla \Psi(\bm{\alpha}_{k-1})- \nabla_{\bm{\alpha}} \nabla_{\bm{x}} \Phi(\bm{\widehat x}_{k-1};\bm{\alpha}_{k-1}) \bm{v}_{k}  \|^2 \nonumber
  \\\leq & \big(1-\frac{\eta\mu_\Phi}{2}+ 4C_vL^2_u\beta^2\big)\big\|  \bm{v}_{k} - ( \nabla^2_{\bm{x}} \Phi(\bm{x}^*_{k-1};\bm{\alpha}_{k-1}) )^{-1}\nabla\widetilde U(\bm{x}_{k-1}^*)  \big\|^2 \nonumber
 \\&+ 2C_v\beta^2 \|\beta^{-1}\big(\bm{\alpha}_{k-1} -  \mathcal{P}_{\mathcal{A}}\big\{ \bm{\alpha}_{k-1} +\beta \nabla \Psi(\bm{\alpha}_{k-1})\big\}\|^2 + \Delta_\Phi + \frac{ 4C_v\beta^2 L^4_u n^2}{\mu^2_\Phi}\delta^2_\Phi. 
\end{align}
Recall  the definition that $G_{\text{proj}}(\bm{\alpha}_k)=\beta^{-1}(  \mathcal{P}_{\mathcal{A}}\{ \bm{\alpha}_{k} +\beta \nabla \Psi(\bm{\alpha}_{k})\}-\bm{\alpha}_{k})$. Then, note that we choose  the stepsize  $\beta$ s.t. $4C_vL^2_u\beta^2<\frac{\eta\mu_\Phi}{4}$. Then, 
telescoping \cref{eq:v_error_final} over $k$ yields
\begin{align*}
\| \bm{v}_{k+1} - ( \nabla^2_{\bm{x}}& \Phi(\bm{x}^*_k;\bm{\alpha}_k) )^{-1}\nabla\widetilde U(\bm{x}_k^*) \|^2 \nonumber
\\\leq& \big(1-\frac{\eta\mu_\Phi}{4}\big)^{k} \| \bm{v}_{1} - ( \nabla^2_{\bm{x}} \Phi(\bm{x}^*_0;\bm{\alpha}_0) )^{-1}\nabla\widetilde U(\bm{x}_0^*) \|^2 \nonumber
\\& +2C_v\beta^2\sum_{t=0}^{k-1}\big(1-\frac{\eta\mu_\Phi}{4}\big)^{k-1-t}\|G_{\text{proj}}(\bm{\alpha}_{t})\|^2 + \frac{4\Delta_\Phi\mu_\Phi+  \eta L^2_u n^2\delta^2_\Phi}{\eta\mu^2_\Phi},
\end{align*}
which, in conjunction with \cref{v:updates}, $\bm{v}_{0}=\bm{0}$ and $\big\|\nabla\widetilde U(\bm{\widehat x}_{0})\big\|\leq \sqrt{n}L_u$, yields 
\begin{align}\label{eq:v_bound_k}
\| \bm{v}_{k+1}& - ( \nabla^2_{\bm{x}} \Phi(\bm{x}^*_k;\bm{\alpha}_k) )^{-1}\nabla\widetilde U(\bm{x}_k^*) \|^2 \leq\big(1-\frac{\eta\mu_\Phi}{4}\big)^{k} 2nL_u^2(1+\mu_\Phi^{-2})   \nonumber
\\& +2C_v\beta^2\sum_{t=0}^{k-1}\big(1-\frac{\eta\mu_\Phi}{4}\big)^{k-1-t}\|G_{\text{proj}}(\bm{\alpha}_{t})\|^2 +\frac{4\Delta_\Phi\mu_\Phi+  \eta L^2_u n^2\delta^2_\Phi}{\eta\mu^2_\Phi}.
\end{align}
Substituting  \cref{eq:v_bound_k} into \cref{eq:mixHest}
\begin{align*}
\big\|\nabla_{\bm{\alpha}} \nabla_{\bm{x}} &\Phi(\bm{\widehat x}_k;\bm{\alpha}_k) \bm{v}_{k+1} - \nabla \Psi(\bm{\alpha}_k) \big\|^2  \nonumber
\\\leq &2L^2_u\big\|\bm{v}_{k+1} - ( \nabla^2_{\bm{x}} \Phi(\bm{x}^*_k;\bm{\alpha}_k) )^{-1}\nabla\widetilde U(\bm{x}_k^*)\big\|^2 + \frac{2n^2L^4_u}{\mu_\Phi^2}\delta^2_\Phi \nonumber
\\\leq &\big(1-\frac{\eta\mu_\Phi}{4}\big)^{k} 4nL_u^4(1+\mu_\Phi^{-2})   +4C_vL_u^2\beta^2\sum_{t=0}^{k-1}\big(1-\frac{\eta\mu_\Phi}{4}\big)^{k-1-t}\|G_{\text{proj}}(\bm{\alpha}_{t})\|^2 \nonumber
\\& +\frac{8L_u^2\Delta_\Phi\mu_\Phi+  4\eta L^4_u n^2\delta^2_\Phi}{\eta\mu^2_\Phi}, 
\end{align*}
which finishes the proof. 

\section{Proof of \Cref{th:mainConvergence}}
Let us first derive the smoothness property of the hypergradient $ \nabla \Psi(\cdot)$. Based on the form of $ \nabla \Psi(\cdot)$ in \cref{eq:hyperGrad}, we have, for any two $\bm{\alpha}_1,\bm{\alpha}_2\in\mathcal{A}$
\begin{align}\label{eq:smoothPhi}
\|\nabla \Psi(&\bm{\alpha}_1) - \nabla \Psi(\bm{\alpha}_2)\| \nonumber
\\\leq& \| \nabla_{\bm{\alpha}} \nabla_{\bm{x}} \Phi(\bm{x}^*(\bm{\alpha}_1);\bm{\alpha}_1) \big( \nabla^2_{\bm{x}} \Phi(\bm{x}^*(\bm{\alpha}_1);\bm{\alpha}_1) \big)^{-1}\nabla\widetilde U(\bm{x}^*(\bm{\alpha}_1)) \nonumber
\\&-\nabla_{\bm{\alpha}} \nabla_{\bm{x}} \Phi(\bm{x}^*(\bm{\alpha}_2);\bm{\alpha}_2) \big( \nabla^2_{\bm{x}} \Phi(\bm{x}^*(\bm{\alpha}_2);\bm{\alpha}_2) \big)^{-1}\nabla\widetilde U(\bm{x}^*(\bm{\alpha}_2))\| \nonumber
\\\overset{(i)}\leq & L_u(\|\bm{x}^*(\bm{\alpha}_1)-\bm{x}^*(\bm{\alpha}_2)\|+\|\bm{\alpha}_1-\bm{\alpha}_2\|) \frac{\sqrt{n}L_u}{\mu_\Phi}  \nonumber
\\&+ \frac{\sqrt{n}L^2_u}{\mu_\Phi^2}\big(L_{\text{Hess}}\|\bm{x}^*(\bm{\alpha}_1)-\bm{x}^*(\bm{\alpha}_2)\|+L_u \|\bm{\alpha}_1-\bm{\alpha}_2\|\big) + \frac{L_u}{\mu_\Phi}\|\bm{x}^*(\bm{\alpha}_1)-\bm{x}^*(\bm{\alpha}_2)\|
\end{align} 
where $(i)$ follows from \Cref{prop:sc}, \Cref{prop:lipschitz_Phi} and \cref{eq:mixHv}. Based on Lemma 2.2 in \cite{ghadimi2018approximation} that $\|\bm{x}^*(\bm{\alpha}_1)-\bm{x}^*(\bm{\alpha}_2)\|\leq \frac{L_{\text{grad}}}{\mu_\Phi}\|\bm{\alpha}_1-\bm{\alpha}_{2}\|$, we obtain from \cref{eq:smoothPhi} that 
\begin{align}\label{eq:smoothnPsiF}
\|\nabla \Psi(\bm{\alpha}_1) - \nabla \Psi(\bm{\alpha}_2)\|\leq \underbrace{\Big(\frac{L_{\text{grad}}}{\mu_\Phi}
 \Big( \frac{\sqrt{n}L_u^2}{\mu_\Phi} + \frac{\sqrt{n}L_u^2L_{\text{Hess}}}{\mu_\Phi^2}+\frac{L_u}{\mu_\Phi}\Big) + \frac{\sqrt{n}L_u^2}{\mu_\Phi} + \frac{\sqrt{n}L_u^3}{\mu_\Phi^2} \Big)}_{L_\Psi}\|\bm{\alpha}_1-\bm{\alpha}_2\|.
\end{align}
Define the hypergradient estimate {\small$\widehat \nabla \Psi(\bm{\alpha}_k)=-\nabla_{\bm{\alpha}}\nabla_{\bm{x}} \Phi(\bm{\widehat x}_k;\bm{\alpha}_k) \bm{v}_{k+1}$} for notational convenience. Then,  
based on the smoothness property established in \cref{eq:smoothnPsiF}, we have 
\begin{align}\label{eq:smoothness_psi}
\Psi(\bm{\alpha}_{k+1}) \geq & \Psi(\bm{\alpha}_{k}) + \langle \nabla\Psi(\bm{\alpha}_{k}), \bm{\alpha}_{k+1}-\bm{\alpha}_{k}\rangle - \frac{L_\Psi}{2}\|\bm{\alpha}_{k+1}-\bm{\alpha}_{k}\|^2 \nonumber
\\\geq &\Psi(\bm{\alpha}_{k}) +  \frac{1}{\beta}\big\langle \beta\widehat \nabla \Psi(\bm{\alpha}_k),  \mathcal{P}_{\mathcal{A}}\big\{\bm{\alpha}_k+\beta \widehat \nabla \Psi(\bm{\alpha}_k) \big\} -\bm{\alpha}_k\big\rangle  \nonumber
\\&+ \big\langle\nabla \Psi(\bm{\alpha}_k)-\widehat \nabla \Psi(\bm{\alpha}_k),\mathcal{P}_{\mathcal{A}}\big\{\bm{\alpha}_k+\beta \widehat \nabla \Psi(\bm{\alpha}_k) \big\} -\bm{\alpha}_k \big\rangle - \frac{L_\Psi}{2}\|\bm{\alpha}_{k+1}-\bm{\alpha}_{k}\|^2. 
\end{align}
For the second term of the right hand side of \cref{eq:smoothness_psi}, we note that 
\begin{align}
\big\langle \beta\widehat \nabla & \Psi(\bm{\alpha}_k),  \mathcal{P}_{\mathcal{A}}\big\{\bm{\alpha}_k+\beta \widehat \nabla \Psi(\bm{\alpha}_k) \big\} -\bm{\alpha}_k\big\rangle  \nonumber
\\=& \big\langle \bm{\alpha}_k+\beta\widehat \nabla  \Psi(\bm{\alpha}_k)-\mathcal{P}_{\mathcal{A}}\big\{\bm{\alpha}_k+\beta \widehat \nabla \Psi(\bm{\alpha}_k) \big\},  \mathcal{P}_{\mathcal{A}}\big\{\bm{\alpha}_k+\beta \widehat \nabla \Psi(\bm{\alpha}_k) \big\} -\bm{\alpha}_k\big\rangle \nonumber
\\& +  \|\bm{\alpha}_k- \mathcal{P}_{\mathcal{A}}\big\{\bm{\alpha}_k+\beta \widehat \nabla \Psi(\bm{\alpha}_k) \big\}\|^2,
\end{align}
which, using the property of the projection on the convex set $\mathcal{A}$, i.e.,  $\langle\bm{x}-\mathcal{P}_{\mathcal{A}}(\bm{x}),\bm{y}-\mathcal{P}_{\mathcal{A}}(\bm{x})\rangle\leq 0$ for any $\bm{y}\in\mathcal{A}$ and noting that $\bm{\alpha}_k=\mathcal{P}_{\mathcal{A}}\{\bm{\alpha}_{k-1}+\beta\widehat \nabla \Psi(\bm{\alpha}_{k-1})\}\in\mathcal{A}$, yields
\begin{align}\label{eq:posiPart}
\big\langle \beta\widehat \nabla & \Psi(\bm{\alpha}_k),  \mathcal{P}_{\mathcal{A}}\big\{\bm{\alpha}_k+\beta \widehat \nabla \Psi(\bm{\alpha}_k) \big\} -\bm{\alpha}_k\big\rangle \geq \|\bm{\alpha}_k- \mathcal{P}_{\mathcal{A}}\big\{\bm{\alpha}_k+\beta \widehat \nabla \Psi(\bm{\alpha}_k) \big\}\|^2.
\end{align}
For notational convenience, let {\small$\widehat G_{\text{proj}}(\bm{\alpha}_k)=\beta^{-1}(\mathcal{P}_{\mathcal{A}}\big\{\bm{\alpha}_k+\beta \widehat \nabla \Psi(\bm{\alpha}_k) \big\}-\bm{\alpha}_k)$} be the estimate of the true generalized projected gradient {\small$G_{\text{proj}}(\bm{\alpha}_k)$} defined in \Cref{prop:hyperGapproximate}. Then,  
substituting \cref{eq:posiPart} into \cref{eq:smoothness_psi} and using that $\langle \bm{a}, \bm{b}\rangle \geq -\frac{1}{2}(\|\bm{a}\|^2 + \|\bm{b}\|^2)$, we have  
\begin{align*}
\Psi(\bm{\alpha}_{k+1}) \geq \Psi(\bm{\alpha}_{k}) + \frac{\beta}{2} \|\widehat G_{\text{proj}}(\bm{\alpha}_k)\|^2 - \frac{\beta}{2}\|\nabla \Psi(\bm{\alpha}_k)-\widehat \nabla \Psi(\bm{\alpha}_k)\|^2- \frac{L_\Psi\beta^2}{2}\|\widehat G_{\text{proj}}(\bm{\alpha}_k)\|^2,
\end{align*}
which, in conjunction with $\|\widehat G_{\text{proj}}(\bm{\alpha}_k)-G_{\text{proj}}(\bm{\alpha}_k)\|\leq \|\nabla \Psi(\bm{\alpha}_k)-\widehat \nabla \Psi(\bm{\alpha}_k)\|$ and $\|\bm{a}+\bm{b}\|^2\geq \frac{1}{2}\|\bm{a}\|^2-\|\bm{b}\|^2$, yields
\begin{align}\label{eq:smoothInterM}
\Psi(\bm{\alpha}_{k+1}) \geq \Psi(\bm{\alpha}_{k}) + \big(\frac{\beta}{4} -\frac{L_\Psi\beta^2}{4}\big)\|G_{\text{proj}}(\bm{\alpha}_k)\|^2 -  \big(\beta -\frac{L_\Psi\beta^2}{2}\big) \|\nabla \Psi(\bm{\alpha}_k)-\widehat \nabla \Psi(\bm{\alpha}_k)\|^2.
 \end{align}
 Applying \Cref{prop:hyperGapproximate} to the above \cref{eq:smoothInterM}, we have  
 \begin{align}\label{eq:lowerGproj}
\Psi(\bm{\alpha}_{k+1}) \geq& \Psi(\bm{\alpha}_{k}) + \big(\frac{\beta}{4} -\frac{L_\Psi\beta^2}{4}\big)\|G_{\text{proj}}(\bm{\alpha}_k)\|^2  \nonumber
\\&-  4C_vL_u^2\beta^2\big(\beta -\frac{L_\Psi\beta^2}{2}\big)\sum_{t=0}^{k-1}\big(1-\frac{\eta\mu_\Phi}{4}\big)^{k-1-t}\|G_{\text{proj}}(\bm{\alpha}_{t})\|^2 \nonumber
\\ &-4nL_u^4(1+\mu_\Phi^{-2})  \big(\beta -\frac{L_\Psi\beta^2}{2}\big)\big(1-\frac{\eta\mu_\Phi}{4}\big)^{k}   - \big(\beta -\frac{L_\Psi\beta^2}{2}\big)\frac{8L_u^2\Delta_\Phi\mu_\Phi+  4\eta L^4_u n^2\delta^2_\Phi}{\eta\mu^2_\Phi}.
 \end{align}
Telescoping \cref{eq:lowerGproj} over $k$ from $0$ to $K-1$ yields
\begin{align*}
\frac{\max_{\bm{\alpha}\in\mathcal{A}}\Psi(\bm{\alpha})-\Psi(\bm{\alpha}_0)}{\beta K} \geq&  \big(\frac{1}{4} -\frac{L_\Psi\beta}{4}\big)\frac{1}{K}\sum_{k=0}^{K-1}\|G_{\text{proj}}(\bm{\alpha}_k)\|^2 \nonumber
\\&- 4\big(1-\frac{L_\Psi\beta}{2}\big)C_vL_u^2\beta^2\frac{1}{K}\sum_{k=1}^{K-1}\sum_{t=0}^{k-1}\big(1-\frac{\eta\mu_\Phi}{4}\big)^{k-1-t}\|G_{\text{proj}}(\bm{\alpha}_{t})\|^2\nonumber
\\&-\frac{16nL_u^4(1+\mu_\Phi^{-2})}{\eta\mu_\Phi}  \big(1 -\frac{L_\Psi\beta}{2}\big)\frac{1}{K} - \big(1 -\frac{L_\Psi\beta}{2}\big)\frac{8L_u^2\Delta_\Phi\mu_\Phi+  4\eta L^4_u n^2\delta^2_\Phi}{\eta\mu^2_\Phi},
\end{align*}
which, in conjunction with $\sum_{k=1}^{K-1}\sum_{t=0}^{k-1}a_{k-1-t}b_t\leq \sum_{k=0}^{K-1}a_k\sum_{t=0}^{K-1}b_t$ for $a_t,b_t\geq 0$, yields
\begin{align}\label{eq:gradNoB}
 \Big(\frac{1}{4} -\frac{L_\Psi\beta}{4}& - 16\big(1-\frac{L_\Psi\beta}{2}\big)\frac{C_vL_u^2\beta^2}{\eta\mu_\Phi}\Big)\frac{1}{K}\sum_{k=0}^{K-1}\|G_{\text{proj}}(\bm{\alpha}_k)\|^2 \leq \frac{\max_{\bm{\alpha}\in\mathcal{A}}\Psi(\bm{\alpha})-\Psi(\bm{\alpha}_0)}{\beta K} \nonumber
\\ &+\frac{16nL_u^4(1+\mu_\Phi^{-2})}{\eta\mu_\Phi}  \big(1 -\frac{L_\Psi\beta}{2}\big)\frac{1}{K} + \big(1 -\frac{L_\Psi\beta}{2}\big)\frac{8L_u^2\Delta_\Phi\mu_\Phi+  4\eta L^4_u n^2\delta^2_\Phi}{\eta\mu^2_\Phi}.
\end{align}
Since we choose $\beta$ such that $0<\beta\leq\frac{1}{2L_\Psi}$, we have $\frac{3}{4}<1-\frac{L_\Psi\beta}{2}<1$. Then, \cref{eq:gradNoB} can be simplified to 
\begin{align}
 \Big(\frac{1}{8}  - \frac{16C_vL_u^2\beta^2}{\eta\mu_\Phi}&\Big)\frac{1}{K}\sum_{k=0}^{K-1}\|G_{\text{proj}}(\bm{\alpha}_k)\|^2 \leq \frac{\max_{\bm{\alpha}\in\mathcal{A}}\Psi(\bm{\alpha})-\Psi(\bm{\alpha}_0)}{\beta K} \nonumber
\\ &+\frac{16nL_u^4(1+\mu_\Phi^{-2})}{\eta\mu_\Phi} \frac{1}{K} +\frac{8L_u^2\Delta_\Phi\mu_\Phi+  4\eta L^4_u n^2\delta^2_\Phi}{\eta\mu^2_\Phi},
\end{align}
which, in conjunction with $\beta\leq\sqrt{\frac{\eta\mu_\Phi}{256C_vL_u^2}}$, yields
\begin{align*}
\frac{1}{K}\sum_{k=0}^{K-1}\|G_{\text{proj}}(\bm{\alpha}_k)\|^2 \leq & \frac{16(\max_{\bm{\alpha}\in\mathcal{A}}\Psi(\bm{\alpha})-\Psi(\bm{\alpha}_0))}{\beta K} \nonumber
\\ &+\frac{256nL_u^4(1+\mu_\Phi^{2})}{\eta\mu_\Phi^3} \frac{1}{K} +\frac{128L_u^2\Delta_\Phi\mu_\Phi+  4\eta L^4_u n^2\delta^2_\Phi}{\eta\mu^2_\Phi},
\end{align*}
which completes the proof. 

\bibliographystyle{unsrt}
\bibliography{refs,inlab-refs}
\end{document}